\documentclass[sigconf,natbib=true,anonymous=false]{acmart}

\AtBeginDocument{%
  \providecommand\BibTeX{{%
    \normalfont B\kern-0.5em{\scshape i\kern-0.25em b}\kern-0.8em\TeX}}}

\setcopyright{acmcopyright}
\copyrightyear{2021}
\acmYear{2021}
\graphicspath{ {Figures/} }
\usepackage[ruled,vlined]{algorithm2e}  % used for algorithm definition
\SetKwInOut{KwInput}{Input} 
\SetKwInOut{KwOutput}{Output}
\usepackage{varwidth}% http://ctan.org/pkg/varwidth
\usepackage{amsfonts}
\usepackage{xspace}
\usepackage{enumitem}
\usepackage{array}
\usepackage{multirow}
\usepackage{caption}
\usepackage{subcaption}
\usepackage{tablefootnote}
\usepackage[export]{adjustbox}
\usepackage{xcolor}
\usepackage{amsthm}

\begin{document}

% \title[short title]{full title}
\title{EqGNN: Equalized Node Opportunity in Graphs}

\author{Uriel Singer}
\email{urielsinger@cs.technion.ac.il}
\orcid{0000-0001-8451-8533}
\affiliation{%
  \institution{Technion, Israel Institute of Technology}
  \city{Haifa}
  \country{Israel}
}
\author{Kira Radinsky}
\email{kirar@cs.technion.ac.il}
\orcid{0000-0001-8451-8533}
\affiliation{%
  \institution{Technion, Israel Institute of Technology}
  \city{Haifa}
  \country{Israel}
}

% \renewcommand{\shortauthors}{Singer et al.}

%\hr{concepts and keywords are not mandatory and we can remove if we are short in space}
\begin{CCSXML}
<ccs2012>
   <concept>
       <concept_id>10010147.10010178</concept_id>
       <concept_desc>Computing methodologies~Artificial intelligence</concept_desc>
       <concept_significance>500</concept_significance>
       </concept>
   <concept>
       <concept_id>10010147.10010257</concept_id>
       <concept_desc>Computing methodologies~Machine learning</concept_desc>
       <concept_significance>500</concept_significance>
       </concept>
 </ccs2012>
\end{CCSXML}

\ccsdesc[500]{Computing methodologies~Artificial intelligence}
\ccsdesc[500]{Computing methodologies~Machine learning}

\keywords{Neural-Networks, Graphs, Graph-Neural-Networks, Fairness}


\newcommand{\specialcell}[2][c]{%
  \begin{tabular}[#1]{@{}c@{}}#2\end{tabular}}

\newcommand{\ig}[1]{\textcolor{blue}{$\ll$\textsf{#1 --YR}$\gg$}}
\newcommand{\us}[1]{\textcolor{red}{$\ll$\textsf{#1 --US}$\gg$}}
\newcommand{\kr}[1]{\textcolor{green}{$\ll$\textsf{#1 --KR}$\gg$}}

\newcommand\footnoteref[1]{\protected@xdef\@thefnmark{\ref{#1}}\@footnotemark}

\newcommand{\ntv}{{node2vec}\xspace}
\newcommand{\Ntv}{{Node2vec}\xspace}
\newcommand{\eqgnn}{{EqGNN}\xspace}
\newcommand{\gcn}{{GCN}\xspace}
\newcommand{\debias}{{Debias}\xspace}
\newcommand{\fairgnn}{{FairGNN}\xspace}

\newcommand{\argmin}{\mathop{\mathrm{argmin}}} 
\def\eqd{\,{\buildrel d \over =}\,} 
%% Tables
\newcolumntype{L}[1]{>{\raggedright\let\newline\\\arraybackslash\hspace{0pt}}m{#1}}
\newcolumntype{C}[1]{>{\centering\let\newline\\\arraybackslash\hspace{0pt}}m{#1}}
\newcolumntype{R}[1]{>{\raggedleft\let\newline\\\arraybackslash\hspace{0pt}}m{#1}}
\newcommand\independent{\protect\mathpalette{\protect\independenT}{\perp}}
\def\independenT#1#2{\mathrel{\rlap{$#1#2$}\mkern2mu{#1#2}}}
\def\multitable#1{\multicolumn{1}{p{0.5cm}}{\centering #1}}

\def\PP{{\mathbb P}}
\def\EE{{\mathbb E}}
\def\RR{{\mathbb R}}
\def\II{{\mathbb I}}
\def\cov{\text{cov}}

\def\g{\mathbf{G}}
\def\F{F}
\def\D{D}
\def\E{\mathbf{E}}
\def\V{\mathbf{V}}
\def\X{\mathbf{X}}
\def\Y{\mathbf{Y}}
\def\A{\mathbf{A}}
\def\y{\mathbf{y}}
\def\a{\mathbf{a}}
\def\B{\mathbf{B}}
\def\h{\mathbf{h}}
\def\P{\mathbf{P}}
\def\Q{\mathbf{Q}}

\newcommand{\change}[1]{\color{red}{#1}\color{black}}
\begin{abstract}
Graph neural networks (GNNs), has been widely used for supervised learning tasks in graphs reaching state-of-the-art results. However, little work was dedicated to creating unbiased GNNs, i.e., where the classification is uncorrelated with sensitive attributes, such as race or gender.
Some ignore the sensitive attributes or optimize for the criteria of statistical parity for fairness. However, it has been shown that neither approaches ensure fairness, but rather cripple the utility of the prediction task.
In this work, we present a GNN framework that allows optimizing representations for the notion of \emph{Equalized Odds} fairness criteria. The architecture is composed of three components: (1) a GNN classifier predicting the utility class, (2) a sampler learning the distribution of the sensitive attributes of the nodes given their labels. It generates samples fed into a (3) discriminator that discriminates between true and sampled sensitive attributes using a novel ``permutation loss'' function. Using these components, we train a model to neglect information regarding the sensitive attribute only with respect to its label. To the best of our knowledge, we are the first to optimize GNNs for the equalized odds criteria. We evaluate our classifier over several graph datasets and sensitive attributes and show our algorithm reaches state-of-the-art results.\footnote{GitHub repository with all code, baselines and data: \url{https://github.com/urielsinger/EqGNN}}
% \footnote{GitHub repository with all code, baselines and data will be available online upon acceptance.}
\end{abstract}

\maketitle

\section{Introduction}
\label{sec:intro}
Supervised learning was shown to exhibit bias depending on the data it was trained on \cite{pedreshi2008discrimination}. This problem is further amplified in graphs, where the graph topology was shown to exhibit different biases \cite{Kang2019ConditionalNE,10.1093/bioinformatics/btaa1036}.
Many popular supervised-learning graph algorithms, such as graph neural networks (GNNs), employ message-passing with features aggregated from neighbors; which might further intensify this bias. 
For example, in social networks, communities are usually more connected between themselves. As GNNs aggregate information from neighbors, it makes it even harder for a classifier to realize the potential of an individual from a discriminated community.

Despite their success \cite{wu2020comprehensive}, little work has been dedicated to creating unbiased GNNs, where the classification is uncorrelated with sensitive attributes, such as race or gender.
The little existing work, focused on ignoring the sensitive attributes \cite{obermeyer2019dissecting}. However,
``fairness through unawareness'' has already been shown to predict sensitive attributes from other features \cite{barocas-hardt-narayanan}. Others \cite{bose2019compositional,dai2020say,buyl2020debayes,rahman2019fairwalk} focused on the criteria of Statistical Parity (SP) for fairness when training node embeddings, which is defined as follows:

\begin{definition}[Statistical parity]
\label{def:sp}
A predictor $\hat{\Y}$ satisfies \emph{statistical parity} with respect to a sensitive attribute $\A$, if $\hat{\Y}$ and $\A$ are independent:
\begin{equation} \label{eq:parity}
\hat{\Y} \independent \A
\end{equation}
\end{definition}

Recently, \cite{dwork2012fairness} showed that SP does not ensure fairness and might actually cripple the utility of the prediction task. 
Consider the target of college acceptance and the sensitive attribute of demographics.
If the target variable correlates with the sensitive attribute, statistical parity would not allow an ideal predictor.
Additionally, the criterion allows accepting qualified applicants in one demographic, but unqualified in another, as long as the percentages of acceptance match.

In recent years, the notion of Equalized Odds (EO) was presented as an alternative fairness criteria \cite{hardt2016equality}. Unlike SP, EO allows dependence on the sensitive attribute $\A$ but only through the target variable $Y$:
\begin{definition}[Equalized odds]
\label{def:eo}
A predictor $\hat{\Y}$ satisfies \emph{equalized odds} with respect to a sensitive attribute $\A$, if $\hat{\Y}$ and $\A$ are independent conditional on the true label $\Y$:
\begin{equation} \label{eq:eqodds}
\hat{\Y} \independent \A \mid \Y
\end{equation}
\end{definition}
The definition encourages the use of features that allow to directly predict $\Y$, while not allowing to leverage $\A$ as a proxy for $\Y$.
Consider our college acceptance example. For the outcome of $\Y$=Accept, we require $\hat{\Y}$ to have similar true and false positive rates across all demographics.
Notice that, $\hat{\Y} = \Y$ aligns with the equalized odds constraint, but we enforce that the accuracy is equally high in all demographics, and penalize models that have good performance on only the majority of demographics.

In this work, we present an architecture that optimizes graph classification for the EO criteria. 
Given a GNN classifier predicting a target class, our architecture expands it with a sampler and a discriminator components.
The goal of the sampler component is to learn the distribution of the sensitive attributes of the nodes given their labels. The sampler generates examples that are then fed into a discriminator. The goal of the latter is to discriminate between true and sampled sensitive attributes. We present a novel loss function the discriminator minimizes -- the \emph{permutation loss}. Unlike cross-entropy loss, that compares two independent or unrelated groups, the permutation loss compares items under two separate scenarios -- with sensitive attribute or with a generated balanced sensitive attribute.

We start by pretraining the sampler, and then train the discriminator along with the GNN classifier using adversarial training.
This joint training allows the model to neglect information regarding the sensitive attribute only with respect to its label, as requested by the equalized odds fairness criteria. 
To the best of our knowledge, our work is the first to optimize GNNs for the equalized odds criteria. 

The contributions of this work are fourfold:
\begin{itemize}
[leftmargin=*,labelindent=1mm,labelsep=1.5mm]
\item We propose \eqgnn, an algorithm with equalized odds regulation for graph classification tasks. 
\item We propose a novel permutation loss which allows us to compare \emph{pairs}. We use this loss in the special case of nodes in two different scenarios -- one under the bias sensitive distribution, and the other under the generated unbiased distribution.
\item We empirically evaluate \eqgnn on several real-world datasets and show superior performance to several baselines both in utility and in bias reduction.
\item We empirically evaluate the permutation loss over both synthetic and real-world datasets and show the importance of leveraging the pair information.
\end{itemize}

\section{Related Work}
\label{sec:rw}
Supervised learning in graphs has been applied in many applications, such as protein-protein interaction prediction \cite{grover16node,singer2019node}, human movement prediction \cite{stgcn2018aaai}, traffic forecasting \cite{yu2018spatio,cui2019traffic} and other urban dynamics \cite{Wang:2017:RRL}.
Many supervised learning algorithms have been suggested for those tasks on graphs, including matrix factorization approaches \cite{Laplacian:NIPS2001,tenenbaum:global:2000,Yan:2007:GEE,roweis2000nonlinear}, random walks approaches \cite{Perozzi:2014:DOL,grover16node} and graph neural network, which recently showed state-of-the-art results on many tasks \cite{wu2020comprehensive}. The latter is an adaptation of neural networks to the graph domain. GNNs create different differential layers that can be added to many different architectures and tasks. GNNs utilize the graph structure by propagating information through the edges and nodes. For instance, GCN~\cite{kipf2017semi} and graphSAGE~\cite{hamilton2017inductive} update the nodes representation by averaging over the representations of all neighbors, while \cite{velivckovic2017graph} proposed an attention mechanism to learn the importance of each specific neighbor.

% The studies over the graph domain in machine learning can be categorized into three categories:  Matrix factorization is the first direction for studying graph representations and includes factorizing the adjacency matrix directly using spectral clustering or PCA \cite{Laplacian:NIPS2001,Roweis:nonlineardimensionality:science,tenenbaum:global:2000,Yan:2007:GEE} or optimizing specifically so it will preserve the edges \cite{ahmed2013distributed}, neighbors \cite{roweis2000nonlinear} or even the k-hop edges \cite{cao2015grarep}. While these approaches are fundamental, newer approaches use random walks on graphs \cite{Perozzi:2014:DOL,grover16node} in order to learn embeddings via neural networks. The newest approach is graph neural networks (GNN) which generalizes neural networks to the graph domain and creates different differential layers that can be added to many different architectures and tasks. The idea behind GNN is to utilize the graph structure by propagating information throw the edges and nodes. For instance, GCN \cite{kipf2017semi} and graphSAGE \cite{hamilton2017inductive} update the nodes representation using all neighbors representation. \cite{velivckovic2017graph} proposed an attention mechanism to learn the importance of each specific edge.

% \subsection{Fairness in Machine Learning}

%\subsection{Fairness in Graphs}
Fairness in graphs was mostly studied in the context of group fairness, by optimizing the SP fairness criteria. 
\cite{rahman2019fairwalk} creates fair random walks by first sampling a sensitive attribute and only then sampling a neighbor from those who hold that specific sensitive attribute. For instance, if most nodes represent men while the minority represent women, the fair random walk promises that the presence of men and women in the random walks will be equal.
\cite{buyl2020debayes} proposed a Bayesian method for learning embeddings by using a biased prior. 
Others, focus on unbiasing the graph prediction task itself rather than the node embeddings. 
For example, \cite{bose2019compositional} uses a set of adversarial filters to remove information about predefined sensitive attributes. It is learned in a self supervised way by using a graph-auto-encoder to reconstruct the graph edges.
\cite{dai2020say} offers a discriminator that discriminates between the nodes sensitive attributes. In their setup, not all nodes sensitive attributes are known, and therefore, they add an additional component that predicts the missing attributes.
\cite{kang2020inform} tackles the challenge of individual fairness in graphs.
In this work, we propose a GNN framework optimizing the EO fairness criteria. 
To the best of our knowledge, our work is the first to study fairness in graphs in the context of EO fairness. 

\section{Equalized-Odds Fair Graph Neural Network}
\label{sec:framework}

\begin{figure*}[ht]
\centering 
	\includegraphics[width=0.9\textwidth]{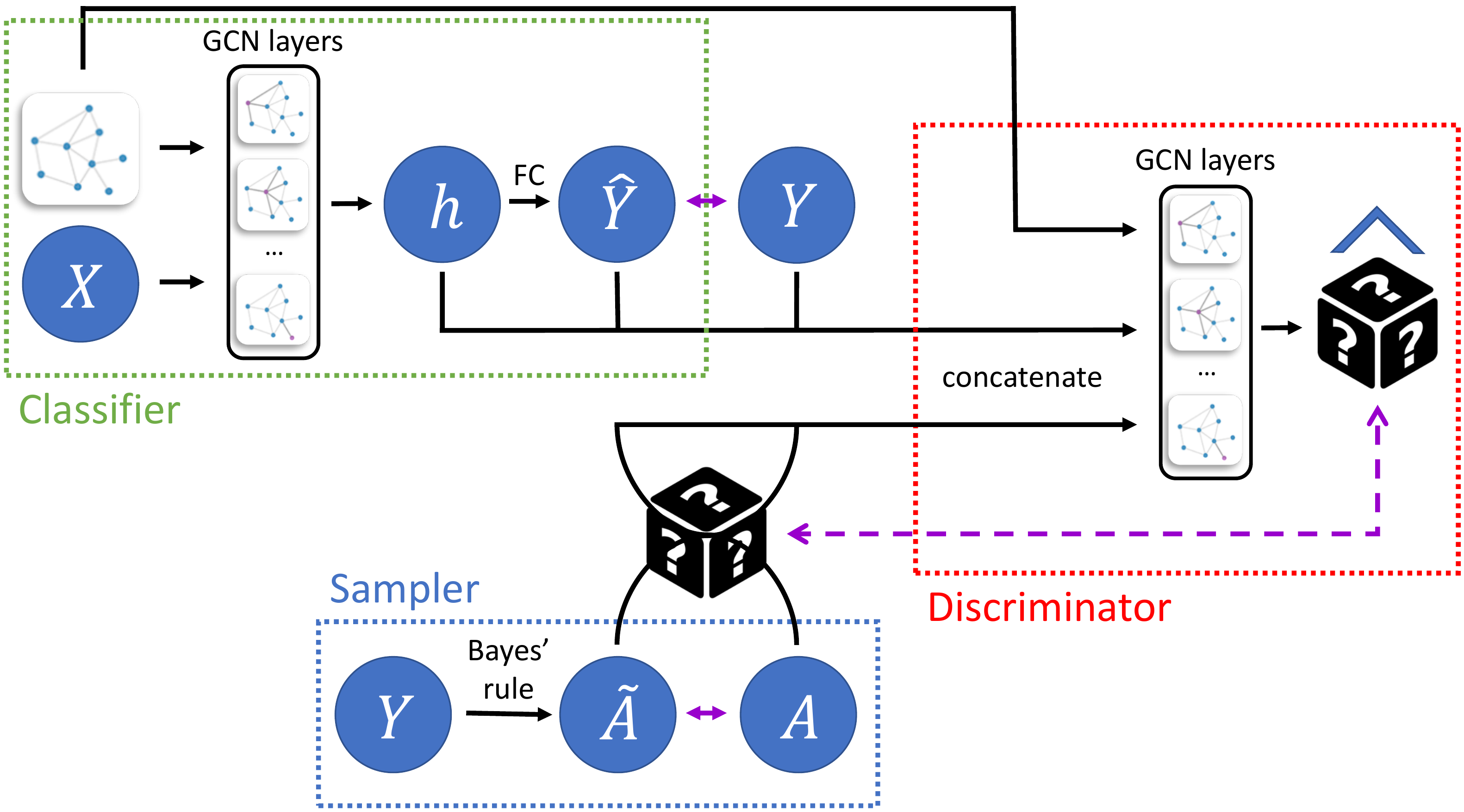}
	\caption{\label{fig:architecture} The full \eqgnn architecture. The blue box represents the sampler model that given a label, samples a \emph{dummy sensitive attribute} (Section~\ref{sec:a_distibution} for details). This model is pretrained independently. The green box represents the classifier, which given a graph and node features, tries to predict its label. The red box represents the discriminator, which minimizes a permutation loss (Section~\ref{sec:permute}). Purple arrows represent loss functions while the magic box represents a random bit ($0$ or $1$), which is used for shuffling the sensitive attribute with its dummy.}
\end{figure*}

Let $\g = (\V, \E)$ be a graph, where $\E$ is the list of edges, $\V$ the list of nodes, $\Y$ the labels, and $\A$ the sensitive attributes.
% Given a node $i$, let $\Y_i  \in \Y$ represent its label and $\A_i \in \A$ its sensitive attribute.
Each node is represented via $n$ features. We denote $X^{|V|\times n}$ as the feature matrix for the nodes in $\V$.
Our goal is to learn a function $\F(\cdot)$ with parameters $\theta_{\F}$, that given a graph $\g$, maps a node $v \in \V$ represented by a feature vector, to its label.

In this work, we present an architecture that can leverage any graph neural network classifier. For simplicity, 
we consider a simple GNN architecture for $\F(\cdot)$ as suggested by \cite{kipf2017semi}: we define $\F(\cdot)$ to be two \gcn~\cite{kipf2017semi} layers, outputting a hidden representation $\h$ for each node. This representation then enters a fully connected layer that outputs $\hat \Y$.
The GNN optimization goal is to minimize the distance between $\hat \Y$ and $\Y$ using a loss function $\ell$. $\ell$ can be categorical cross-entropy (CCE) for multi-class classification, binary cross-entropy (BCE) for binary classification, or mean square error (L2) for regression problems.
In this work, we extend the optimization to $\min_{\theta_{\F}} L_{task} = \ell(\hat{\Y},\Y)$ while satisfying Eq. \ref{eq:eqodds} for fair prediction.  
% \begin{equation}
% \min_{\theta_{f}} L_{task} = \ell(\hat{\Y},\Y),
% \end{equation}
%where $\ell$ is one of CCE, BCE, L2 depending on the problem.

We propose a method, \eqgnn, that trains a GNN model to neglect information regarding the sensitive attribute only with respect to its label. 
The full architecture of \eqgnn is depicted in Figure~\ref{fig:architecture}.
Our method pretrains a sampler (Section~\ref{sec:sampler}), to learn the distribution $\PP_{\A \mid \Y}$, of the sensitive attributes of the nodes given their labels (marked in blue in Figure~\ref{fig:architecture}).
We train a GNN classifier (marked in green in Figure~\ref{fig:architecture}), while regularizing it with a discriminator (marked in red in Figure~\ref{fig:architecture}) that discriminates between true and sampled sensitive attributes.
Section~\ref{sec:discriminator} presents the EO regulation. The regularization is done using a novel loss function -- the ``permutation loss'', which is capable of comparing paired samples (formally presented in Section~\ref{sec:permute}, and implementation details are discussed in Section~\ref{sec:permute_disc}).
For the unique setup of adversarial learning over graphs, we show that incorporating the permutation loss in a discriminator, brings performance gains both in utility and in EO. 
Section~\ref{sec:full} presents the full \eqgnn model optimization procedure.

\subsection{Sampler}
\label{sec:sampler}
% Before we train our generative model (combined from the classifier and discriminator), we start by training a model that will be used for sampling `negatives` while training the discriminator. 
To comply with the SP criteria (Eq. \ref{eq:parity}), given a sample $i$, we wish the prediction of the classifier, $\hat{\Y}_i$, to be independent of the sample's sensitive attribute $\A_i$. In order to check if this criteria is kept, we can sample a fake attribute out of $\PP_{\A}$ (e.g., in case of equal sized groups, a random attribute), and check if $\hat{\Y}_i$ can predict the true or fake attribute. If it is not able to predict, this means that $\hat{\Y}_i$ is independent of $\A_i$ and the SP criteria is kept. As all information of $\hat{\Y}_i$ is also represented in the hidden representation $\h_i$, one can simply train a discriminator to predict the sensitive attribute given the hidden representation $\h_i$. A similar idea was suggested by \cite{dai2020say,bose2019compositional}.

% Alternatively, to comply with the EO criteria (Eq. \ref{eq:eqodds}), 

To comply with the EO criteria (Eq. \ref{eq:eqodds}), the classifier should not be able to separate between an example with a real sensitive attribute $\A_i$ and an example with an attribute sampled from the conditional distribution $\PP_{\A \mid \Y}(\A_i \mid \Y_i)$. Therefore, we jointly train the classifier with a discriminator that learns to separate between the two examples. 
Formally, given a sample $i$, we would want the prediction of the classifier, $\hat{\Y}_i$, to be independent of the attribute $\A_i$ only given its label $\Y_i$. Thus, instead of sampling the fake attribute out of $\PP_{\A}$ we sample the fake attribute out of $\PP_{\A \mid \Y}$. 

We continue describing the sensitive attribute distribution learning process, $\PP_{\A \mid \Y}$, and then present how the model samples \emph{dummy attributes} that will be used as ``negative'' examples for the discriminator.

\subsubsection{Sensitive Attribute Distribution Learning}
\label{sec:a_distibution}
% \cite{romano2020achieving} showed that by observing Eq.~\ref{eq:eqodds} we understand that learning the distribution $\PP_{\A|\Y}$ is necessary for training the classifier.
Here, our goal is to learn the distribution $\PP_{\A \mid \Y}$.
For a specific sensitive attribute $a$ and label $y$, the probability can be expressed using Bayes' rule as: 

\begin{equation}
\begin{split}
\label{eq:sampling_dummies}
\PP_{\A \mid \Y}(\A=a \mid \Y=y) = \frac{\PP(\Y=y \mid \A=a)\PP(\A=a)}{\sum_{a' \in \A} \PP(\Y=y \mid \A=a')\PP(\A=a')}
\end{split}
\end{equation}
     
The term $\PP(\Y=y \mid \A=a)$ can be derived from the data by counting the number of samples that are both with label $y$ and sensitive attribute $a$, divided by the number of samples with sensitive attribute $a$. Similarly, $\PP(\A=a)$ is calculated as the number of samples with sensitive attribute $a$, divided by the total number of samples. In a regression setup, these can be approximated using a linear kernel density estimation.
% \us{We additionally experimented with a GNN model predicting $\A$ given $\Y$, and the graph $\g$. This resulted with inferior performance, and therefore left the the sampler to be learned using Bayes' rule.}

\subsubsection{Fair Dummy Attributes}
\label{sec:dummy}
% Having a model $\PP_{\A|\Y}$ that estimates the equalized odds distribution of the sensitive attributes is not enough. This model should be utilised for creating ``negative'' samples for the discriminator.
During training of the end-to-end model (Section~\ref{sec:full}), the sampler receives a training example $(\X_i, \Y_i, \A_i)$
and generates a \emph{dummy attribute} by sampling $\tilde{\A}_i \sim \PP_{\A \mid \Y}(\A_i \mid \Y_i)$.
Notice that $\A_i$ and $\tilde{\A}_i$ are equally distributed given $\Y_i$. This ensures that if the classifier holds the EO criteria, then $(\X_i, \Y_i, \A_i)$ and $(\X_i, \Y_i, \tilde{\A}_i)$ will receive an identical classification, whereas otherwise it will result in different classifications. In Section~\ref{sec:discriminator} we further explain the optimization process that utilizes the \emph{dummy attributes} for regularizing the classifier for EO.

% \subsection{Classifier}
% \label{sec:classifer}
% Given a sample $(\X_i, \Y_i, \A_i)$ our goal is to learn a function $f(\cdot)$ with parameters $\theta_{f}$ that maps $\X_i$ to $\Y_i$. We define $f(\cdot)$ to be a 2-layered \gcn \cite{kipf2017semi} outputting a hidden layer $\h_i$, a representation for each node. This representation then enters a fully connected layer that outputs $\hat \Y_i$. This way we have both a prediction and a representation for each node. While the representation will be used only for the discriminator, the prediction $\hat \Y_i$ is then compared with the true $\Y_i$ using categorical cross entropy for multi class classification creating the loss function:
% \begin{equation}
% \min_{\theta_{f}} L_{task} = CCE(\hat{\Y},\Y) = -\sum_{i}\sum_{\Y_i^k \in \Y_i}\Y_i^k\log(\hat{\Y}_i^k)
% \end{equation}
% where for binary classification we use the binary cross entropy loss:
% \begin{equation}
% \begin{split}
% \min_{\theta_{f}} L_{task} = BCE(\hat{\Y},\Y)  = \\
% -\sum_{i}(\Y_i\log(\hat{\Y}_i) + (1-\Y_i)\log(1-\hat{\Y}_i))
% \end{split}
% \end{equation}
% and for regression problems we use the mean square error loss:
% \begin{equation}
% \min_{\theta_{f}} L_{task} = L_2(\hat{\Y},\Y)  = \sum_{i}(\Y_i-\hat{\Y}_i)^2
% \end{equation}

% The architecture presented in this paper can use any of the known graph neural network approaches. Specifically, we opted to work with \gcn~\cite{kipf2017semi}, which achieved state-of-the-art performance on various benchmarks.

\subsection{Discriminator}
\label{sec:discriminator}
A GNN classifier without regulation might learn to predict biased node labels based on their sensitive attributes.
To satisfy EO, the classifier should be unable to distinguish between real examples and generated examples with \emph{dummy attributes}. Therefore, we utilize an adversarial learning process and add a discriminator that learns to distinguish between real and fake examples with \emph{dummy attributes}. Intuitively, this regularizes the classifier to comply with the EO criterion.

% Observing again equation (\ref{eq:eqodds}), we see that we should define a sample to hold the triplet $(\Y_i,\A_i, \hat{\Y}_i)$. In order to add more signal to the problem we add the hidden representation $\h_i$ we learned via $f(\cdot)$. By concatenating the four $(\Y_i, \A_i, \hat{\Y}_i, \h_i)$ we create a true sample for the discriminator. We then sample a \emph{dummy attribute} using the Sampler each time the discriminator is called: $\tilde{\A}_i \sim \PP_{\A|\Y}(\A_i|\Y_i)$. By concatenating the four $(\Y_i, \tilde{\A}_i, \hat{\Y}_i, \h_i)$ we create the negative sample.
% Given a sample (true or fake) the function $d(\cdot)$ 
Intuitively, one might consider $\h_i$, the last hidden layer of the classifier $\F(\cdot)$, as the unbiased representation of node $i$.
The discriminator receives two types of examples: (1) real examples $(\Y_i, \A_i, \hat{\Y}_i,\h_i)$ and (2) negative examples $(\Y_i, \tilde{\A}_i, \hat{\Y}_i, \h_i)$, where $\tilde{\A}_i \sim \PP_{\A \mid \Y}(\A_i \mid \Y_i)$ is generated by the pretrained sampler.
The discriminator learns a function $\D(\cdot)$ with parameters $\theta_{\D}$, that given a sample, classifies it to be true or fake. The classifier in its turn tries to ``fool'' it. This ensures the classifier doesn't hold bias towards specific labels and sensitive attributes and keeps the EO criterion.
The formal adversarial loss is defined as:
\begin{equation}
\label{eq:adv_classifier}
\begin{split}
\min_{\theta_{\F}}\max_{\theta_{\D}} L_{adv} = \EE \left[ log(\D(\Y  \; || \; \A \; || \; \F(\X,\g) \; || \; \h)) \right] \\
\qquad + \EE_{\tilde{\A} \sim \PP_{\A \mid \Y}(\A \mid \Y)} \left[ log(1-\D(\Y  \; || \; \tilde{\A} \; || \; \F(\X,\g) \; || \; \h)) \right]
\end{split}
\end{equation}
where $\EE$ is the expected value, and $||$ represents the concatenation operator. The discriminator tries to maximize $L_{adv}$ to distinguish between true and fake samples, while the classifier tries to minimize it, in order to ``fool'' him.
In our implementation, $\D(\cdot)$ is a GNN with two \gcn layers outputting the probability of being a true sample.
% As the graph structure may hold signal regarding fake and true samples, we formalize $d(\cdot)$ by a similar architecture to $f(\cdot)$: two \gcn layers followed by a fully connected layer outputting the probability being a true sample. As $d(\cdot)$ is a graph neural network, all real samples enter $d(\cdot)$ together followed by all fake samples.

we observe that true and fake attributes are paired (the same node, with a real sensitive attribute or a \emph{dummy attribute}).
A binary loss as defined in Eq.~\ref{eq:adv_classifier} holds for unpaired examples, and does not take advantage of knowing the fake and true attributes are paired. Therefore, we upgrade the loss to handle paired nodes by utilizing the permutation loss. We first formally define and explain the permutation loss in Section~\ref{sec:permute}, and then continue discussing its implantation details in our architecture in Section~\ref{sec:permute_disc}.

\subsection{Permutation Loss}
\label{sec:permute}
In this section, we formally define the new permutation loss test presented in this work.
Let us assume $\X_1$ and $\X_2$ are two groups of subjects. Many applications are interested of learning and understanding the difference between the two. For example, in the case where $\X_1$ represents test results of students in one class, while $\X_2$ represents test results of a different class, it will be interesting to check if the two classes are equally distributed (i.e., $\PP(\X_1) \sim \PP(\X_2)$).

\begin{definition}[T-test]
\label{def:ttest}
Given two groups $\X_1, \X_2 \subset \RR$, the statistical difference can be measured by the t-statistic:
$$t=\frac {{\bar{\X}_1-\bar{\X}}_2}{\sqrt {\frac{s_1^2}{n_1}+\frac {s_2^2}{n_2}}}$$
Where $\bar{\X}_i$, $s_i$, and $n_i$ are the means, variances, and group sizes respectively.
\end{definition}

While this test assumes $\X_1, \X_2$ are scalars that are normally distributed, \cite[]{lopez2016revisiting} proposed a method called C2ST that handles cases where $\X_1, \X_2 \subset \RR^d$ for $d \geq 1$. They proposed a classifier that is trained to predict the correct group of a given sample, which belongs to either $X_1$ or $X_2$. By doing so, given a test-set, they are able to calculate the t-statistic by simply checking the number of correct predictions:

\begin{definition}[C2ST]
\label{def:c2st}
Given two groups $\X_1, \X_2 \subset \RR^d$ (labeled 0 and 1 respectively), the statistical difference can be measured by the t-statistic \cite{lopez2016revisiting}:
$$t=\frac{1}{N} \sum^{N}_{i=1}{\II \left[ \II \left( f(x_i)>\frac{1}{2} \right) =y_i \right] }$$
where $x_i \in \X_1 \cup \X_2$, $y_i$ is $x_i$'s original group label, $N=|\X_1| + |\X_2|$ is the number of samples in the test-set, and $f$ is a trained classifier outputting the probability a sample is sampled from group $1$.
\end{definition}

The basic (and yet important) idea behind this test is that if a classifier is not able to predict the group of a given sample, then the groups are equality distributed. Mathematically, the t-statistic can be calculated by the number of correct samples of the classifier.

However, the C2ST criterion is not optimal when the samples are paired ($\X_1^i$ and $\X_2^i$ represent the same subject), as it doesn't leverage the paired information. Consider the following example: assuming the pairs follow the following rule: $\X_1 \sim \mathcal{N}_d(\textbf{0},\textbf{1})$ and $\X^i_2=\X^i_1+\epsilon \quad \forall \X^i_2 \in \X_2$. As the two Gaussians overlap, a simple linear classifier will not be able to detect any significant difference between the two groups, while we hold the information that there is a significant difference ($\X_1^i$ is always smaller in $\epsilon$ from its pair $\X_2^i$). Therefore, it is necessary to define a new test that can manage paired examples.

\begin{definition}[Paired T-test]
\label{def:pairedttest}
Given two paired groups $\X_1, \X_2 \subset \RR$, and $\X_D=\X_1-\X_2$, the statistical difference can be measured by the t-statistic:
$$t=\frac{\bar{\X}_D}{\frac{s_D}{\sqrt{n}}}$$
where $\bar{\X}_D$, $s_D$ are the mean and standard deviation of $\X_D$, and $n=|\X_D|$ is the number of pairs.
\end{definition}

Again, this Paired T-test assumes $\X_1, \X_2$ are scalars that are normally distributed. A naive adaptation of \cite{lopez2016revisiting} for paired data with $d \geq  1$, would be to first map the pairs into scalars and then calculate their differences (as known in the paired student t-test for $d=1$). This approach also assumes the samples are normally distributed, and therefore is not robust enough. An alternative to the paired t-test is the permutation test \cite{oden1975arguments}, which has no assumptions on the distribution. It checks how different the t-statistic of a specific permutation is from many (or all) other random permutations t-statistics. By doing so, it is able to calculate the p-value of that specific permutation. We suggest a differential version of the permutation test. This is done by using a neural network architecture that receives either a real permutation or a shuffled, and tries to predict the true permutation.

\begin{algorithm}[ht]
\caption{Permutation Loss}
\label{alg:permute}
	\KwInput{$\X_1$, $\X_2$ - paired groups in $\RR^d$ \\
		     $lr$ - learning rate}
	\KwOutput{t - the t-statistic}
    \nl $\X^{train}_1$, $\X^{train}_2$, $\X^{test}_1$, $\X^{test}_2$ = train\_test\_split($\X_1$, $\X_2$) \\
    \nl $N_{train}, N_{test} = |\X^{train}_1|, |\X^{test}_1|$ \\
	\nl	\While{$f$ not converged}{
	\nl             $l = random([0,1], size=N_{train})$ \\
	\nl             $pairs = concat([\X^{train}_1, \X^{train}_2])$ \\
	\nl             $shuffled\_pairs = concat([pairs[true\_permute],$ \\ \qquad\qquad\qquad\qquad\qquad\quad $pairs[1-true\_permute]])$ \\
	\nl             $\hat{l} = f(shuffled\_pairs)$  \\
	\nl             $Loss=BinaryCrossEntropy(\hat{l}, l)$ \\
	\nl             $f \leftarrow f-lr \cdot \nabla_f Loss$  \\
	}
	\nl $test\_pairs = concat([\X^{test}_1, \X^{test}_2])$ \\
	\nl $l = random([0,1], size=N_{test})$ \\
	\nl $\hat{l} = f(test\_pairs)$  \\
	\nl $t = \frac{1}{N_{test}} \sum^{N_{test}}_{i=1}{\II \left[ \II \left( \hat{l_i}>\frac{1}{2} \right) = l_i \right] }$ \\
	\nl return $t$
\end{algorithm}

In Algorithm \ref{alg:permute}, we define the full differential permutation loss. The permutation phase consists of four steps: (1) For each pair, $\X^i_1, \X^i_2$ each of size $d$, sample a random number $l_i \in \{0,1\}$ (line 4 in the algorithm). (2) If $l_i=0$, concatenate the pair in the original order: $[\X^i_1, \X^i_2]$, while if $l_i=1$, concatenate in the permuted order: $[\X^i_2, \X^i_1]$. This resolves us with a vector of size $2d$ (lines 5-6). (3) This sample enters a classifier that tries to predict $l_i$ using binary-cross-entropy (lines 7-8). (4) We update the classifier weights (line 9), and return to step 1, until convergence. Assuming $\X_1$ and $\X_2$ are exchangeable, the classifier will not be able to distinguish if a permutation was performed or not.
The idea behind this test is that if a classifier is not able to predict the true ordering of the pairs, it means that there is no significant difference between this specific permutation or any other permutation. Mathematically, similarly to C2ST, the t-statistic can be calculated by the number of correct samples of the classifier.

While this is similar to the motivation of the naive permutation test \cite{oden1975arguments}, we offer additional benefits: (1) The test is differential, meaning we can represent it using a neural network as a layer (see Section~\ref{sec:permute_disc}). (2) The test can handle $d \geq  1$. (3) We do not need to define a specific t-statistic for each problem, but rather the classifier checks for any possible signal.

As a real life example that explains the power of the loss, assume a person is given a real coin and fake coin. While she observes each one separately, her confidence of which is which, will be much less if she would rather receive them together. This real life example demonstrates the important difference between C2ST and the permutation loss (see  Section~\ref{sec:result_premute_synthetic} for an additional example over synthetic data).

\subsection{Permutation Discriminator}
\label{sec:permute_disc}
Going back to our discriminator, we observe that true and fake attributes are paired (the same node, with a real sensitive attribute or a \emph{dummy attribute}).
% A simple binary loss as defined in Eq.~\ref{eq:adv_classifier} holds for unpaired examples, therefore we upgrade the loss to handle paired nodes by utilizing the permutation loss we defined in Section ~\ref{sec:permute}.
We create paired samples: $(\Y_i, \A_i, \tilde{\A}_i, \hat{\Y}_i, \h_i)$. At each step we randomly permute the sensitive attribute and its \emph{dummy attribute}, creating a sample labeled as permuted or not. Now our samples are  $(\Y_i, \A_i, \tilde{\A}_i, \hat{\Y}_i, \h_i)$ with label $0$ indicating no permutation was applied, while $(\Y_i, \tilde{\A}_i, \A_i, \hat{\Y}_i, \h_i)$ with label $1$ indicating a permutation was applied. The discriminator therefore receives the samples and predicts the probability of whether a permutation was applied. We therefore adapt the adversarial loss of Eq.~\ref{eq:adv_classifier} to:
\begin{equation}
\label{eq:adv_permute}
\begin{split}
% \min_{\theta_{\F}}\max_{\theta_{\D}} L_{adv} = \EE_{s \sim (\Y, \tilde{\A}, \A, \F(\X,\g), \h)}[log(\D(s)] \\
% + \EE_{s \sim (Y, \A, \tilde{\A}, \F(\X,\g), \h)}[log(1-\D(s)]
\min_{\theta_{\F}}\max_{\theta_{\D}} L_{adv} = \EE_{\tilde{\A} \sim \PP_{\A \mid \Y}(\A \mid \Y)} \left[ log(\D(\Y  \; || \; \tilde{\A} \; || \; \A \; || \; \F(\X,\g) \; || \; \h)) \right] \\
\qquad + \EE_{\tilde{\A} \sim \PP_{\A \mid \Y}(\A \mid \Y)} \left[ log(1-\D(\Y  \; || \; \A \; || \; \tilde{\A} \; || \; \F(\X,\g) \; || \; \h)) \right]
\end{split}
\end{equation}
The loss used in the permutation test is a binary cross-entropy and therefore convex.

As an additional final regulation and to improve stability of the classifier, similarly to \cite{romano2020achieving}, we propose to minimize the absolute difference between the covariance of $\hat{\Y}$ and $\A$ from the covariance of $\hat{\Y}$ and $\tilde{\A}$:

\begin{equation}
\min_{\theta_{\F}} L_{cov}= \|cov(\hat{\Y}, \A) - cov(\hat{\Y}, \tilde{\A})\|^2
\end{equation}

\subsection{\eqgnn}
\label{sec:full}
The sampler is pretrained using Eq.~\ref{eq:sampling_dummies}.
We then jointly train the classifier and the discriminator optimizing the objective function:
\begin{equation}
\label{eq:final}
\begin{split}
\min_{\theta_{\F}}\max_{\theta_{\D}}  L_{task} + \lambda(L_{adv} + \gamma L_{cov}),
\end{split}
\end{equation}
where $\theta_{\F}$ are the parameters of the classifier and $\theta_{\D}$ are the parameters of the discriminator. $\lambda$ and $\gamma$ are hyper-parameters that are used to tune the different regulations. This objective is then optimized for $\theta_{\F}$ and $\theta_{\D}$ one step at a time using the Adam optimizer \cite{kingma2014adam}, with learning rate $10^{-3}$ and weight-decay $10^{-5}$.
The training is further detailed in Algorithm \ref{alg:full}.

\begin{algorithm}[ht]
\caption{\eqgnn training procedure}
\label{alg:full}
	\KwInput{$\X$ - node features \\
	        $\g$ - graph \\
	        $\A$ - node sensitive attributes \\
	        $\Y$ - node labels \\
	        $N$ - number of nodes \\
	        $lr$ - learning rate}
	\KwOutput{$\F$ - best according to $val\_loss$}
    \nl $Sampler = \PP_{\A \mid \Y}$ \text{\qquad // calculate using equation (\ref{eq:sampling_dummies})} \\
	\nl	\While{$\F$ not converged}{
	\nl             $\hat{\Y}, \h = \F(\X, \g)$ \\
	\nl             $\tilde{\A} \sim Sampler(\A \mid \Y)$ \text{ // sample \emph{dummy sensitive attributes}}\\
	\nl             $L_{task} = \ell(\hat{\Y}, \Y)$ \\
	\nl             $l = random([0,1], size=N)$ \\
	\nl             $attrs = [\A, \tilde{\A}]$ \\
	\nl             $shuffled\_pairs = concat([\Y, attrs[true\_permute],$ \\ \qquad\qquad\qquad\qquad\qquad $attrs[1-true\_permute], \hat{\Y}, \h])$ \\
	\nl             $\hat{l} = \D(shuffled\_pairs, \g)$  \\
	\nl             $L^F_{adv} = BinaryCrossEntropy(\hat{l}, 1-l)$ \\
	\nl             $L_{cov}= \|cov(\hat{\Y}, \A) - cov(\hat{\Y}, \tilde{\A})\|^2$ \\
	\nl             $\F \leftarrow \F-lr \cdot \nabla_\F(L_{task} + \lambda(L^\F_{adv} + \gamma L_{cov}))$  \\
	\nl             $L^\D_{adv} = BinaryCrossEntropy(\hat{l}, l)$  \\
	\nl             $\D \leftarrow \D-lr \cdot \nabla_\D L^\D_{adv}$ \\
	}
	\nl return $\F$
\end{algorithm}

\section{Experimental Setup}
\label{sec:experiments}

In this section, we describe our datasets, baselines, and metrics. Our baselines include fair baselines designed specifically for graphs, and general fair baselines that we adapted to the graph domain.

\subsection{Datasets}
Table~\ref{table:datasets} summarizes the datasets' characteristics used for our experiments. Intra-group edges are the edges between similar sensitive attributes, while inter-group edges are edges between different sensitive attributes.

\textbf{Pokec}~\cite{takac2012data}.
Pokec is a popular social network in Slovakia. An anonymized snapshot of the network was taken in 2012. User profiles include gender, age, hobbies, interest, education, etc. The original Pokec dataset contains millions of users. We sampled a sub-network of the ``Zilinsky`` province. 
We create two datasets, where the sensitive attribute in one is the gender, and region in the other.  
The label used for classification is the job of the user. The job field was grouped in the following way: (1)``education`` and ``student``, (2)``services \& trade`` and ``construction``, and (3) ``unemployed``. 
% While we want the network signal of users with no labels, we are not interested in using them as labeled data, therefore, we labeled them as a fourth group while masking them out during training and testing.

\textbf{NBA}~\cite{dai2020say}
This dataset was presented in the \fairgnn baseline paper. The NBA Kaggle dataset contains around 400 basketball players with features including performance statistics, nationality, age, etc. This dataset was extended in \cite{dai2020say} to include the relationships of the NBA basketball players on Twitter. The binary sensitive attribute is whether a player is a U.S. player or an overseas player, while the task is to predict whether a salary of the player is over the median.

\begin{table}[tb]
\caption{\label{table:datasets} Datasets' characteristics.}
\centering\small
\setlength{\tabcolsep}{0.5em}
\begin{tabular}{cccc}
\toprule
Dataset                 & Pokec-region  & Pokec-gender & NBA          \\
\midrule
\# of nodes             & $67,796$      & $67,796$     & $355$        \\
\# of attributes        & $276$         & $276$        & $95$         \\
\# of edges             & $617,958$     & $617,958$    & $9,477$      \\
sensitive groups ratio  & $1.84$        & $1.02$       & $3.08$       \\
\# of inter-group edges & $30,519$      & $339,461$    & $2,472$      \\
\# of intra-group edges & $587,439$     & $278,497$    & $7,005$      \\
\bottomrule
\end{tabular}
\end{table}

\subsection{Baselines}
In our evaluation, we compare to the following baselines:

\textbf{\gcn}~\cite{kipf2017semi}: \gcn is a classic GNN layer that updates a node representation by averaging the representations of his neighbors. For fair comparison, we implemented \gcn as the classifier of the \eqgnn architecture (i.e., an unregulated baseline, with the only difference of $\lambda=0$).

\textbf{\debias}~\cite{zhang2018mitigating}: %\debias is a method that proposes different implementations for optimizing EO or SP. We chose to use the EO architecture. The idea
\debias optimizes EO by using a discriminator that given $\Y$ and $\tilde{\Y}$ predicts the sensitive attribute. While \debias is a non-graph architecture, for fair comparison, we implemented \debias with the exact architecture as \eqgnn. Unlike \eqgnn, \debias's  discriminator receives as input only $Y$ and $\hat{Y}$ (without the sensitive attribute or \emph{dummy attribute}) and predicts the sensitive attribute. As the discriminator receives $\Y$, it neglects the sensitive information with respect to $\Y$ and, therefore optimizes for EO.

\textbf{\fairgnn}~\cite{dai2020say}: \fairgnn uses a discriminator that, given $\h$, predicts the sensitive attribute. By doing so, they neglect the sensitive information from $\h$. As this is without respect to $\Y$, they optimize SP (further explained in Section~\ref{sec:sampler}). \fairgnn offers an additional predictor for nodes with unknown sensitive attributes. As our setup includes all nodes' sensitive attributes, this predictor is irrelevant. We opted to use FairGCN for fair comparison. In addition, we generalized their architecture to support multi-class classification.

For all baselines, $50\%$ of nodes are used for training, $25\%$ for validation and $25\%$ for testing. The validation set is used for choosing the best model for each baseline throughout the training. As the classifier is the only part of the architecture used for testing, an early stopping was implemented after its validation loss (Eq.~\ref{eq:final}) hasn't improved for $50$ epochs. The epoch with the best validation loss was then used for testing. All results are averaged over $20$ different train/validation/test splits for Pokec datasets and $40$ for the NBA dataset.
For fair comparison, we implemented grid-search for all baselines over $\lambda \in \{0.01, 0.1, 1, 10\}$ for baselines with a discriminator, and $\gamma \in \{0, 50\}$ for baselines with a covariance expression. For both Pokec datasets and for all baselines $\lambda=1$ and $\gamma=50$, while for NBA we end up using $\lambda=0.1$ and $\gamma=50$ expect for \fairgnn with $\lambda=0.01$.
All experiments used a single Nvidia P100 GPU with the average run of 5 minutes per seed for Pokec and 1 minute for NBA. 
Results where logged and analyzed using the \cite{wandb,falcon2019pytorch} platforms.

\subsection{Metrics}
\label{sec:metrics}

\subsubsection{Fairness Metrics}
\paragraph{Equalized odds.}
The definition in Eq.~\ref{eq:eqodds}, can be formally written as:
\begin{equation}
\begin{split}
\label{eq:eqodds_proba}
\PP(\hat{\Y}=y|\Y=y,\A=a_1) = \PP(\hat{\Y}=y|\Y=y,\A=a_2) \\ \forall a_1,a_2 \in \A, \forall y \in \Y
\end{split}
\end{equation}
The value of $\PP(\hat{\Y}=y|\Y=y,\A=a)$ can be calculated from the test-set as follows: given all samples with label $y$ and sensitive attribute $a$, we calculate the proportion of samples that where labeled $y$ by the model.
As we handle a binary sensitive attribute, given a label $y \in \Y$, we calculate the absolute difference between the two sensitive attribute values:

\begin{equation}
\begin{split}
\label{eq:eqodds_metrics_class}
\Delta EO(y) = |\PP(\hat{\Y}=y|\Y=y,\A=0) - \PP(\hat{\Y}=y|\Y=y,\A=1)|
\end{split}
\end{equation}

According to Eq. \ref{eq:eqodds_proba}, our goal is to have both probabilities equal. Therefore, we desire $\Delta EO(y)$ to strive to $0$. 
We finally aggregate $\Delta EO(y)$ for all labels using the max operator to get a final scalar metric:

\begin{equation}
\begin{split}
\label{eq:eqodds_metrics}
\Delta EO = max(\{\Delta EO(y) | y \in \Y\})
\end{split}
\end{equation}

As we propose an equalized odds architecture, $\Delta EO$ is our main fairness metric.

\paragraph{Statistical parity.}
The definition in Eq.~\ref{eq:parity} can be formally written as:

\begin{equation}
\begin{split}
\label{eq:parity_proba}
\PP(\hat{\Y}=y|\A=a_1) = \PP(\hat{\Y}=y|\A=a_2) \quad \forall a_1,a_2 \in \A
\end{split}
\end{equation}
The value of $\PP(\hat{\Y}=y|\A=a)$ can be calculated from the test-set the following way: given all samples with sensitive attribute $a$, we calculate the proportion of samples that where labeled $y$ by the model.
As we handle a binary sensitive attribute, given a label $y \in \Y$ we calculate the absolute difference between the two sensitive attribute values:

\begin{equation}
\begin{split}
\label{eq:parity_metrics_class}
\Delta SP(y) = |\PP(\hat{\Y}=y|\A=0) - \PP(\hat{\Y}=y|\A=1)|
\end{split}
\end{equation}
According to Eq. \ref{eq:parity_proba}, our goal is to have both probabilities equal. Therefore, we desire $\Delta SP(y)$ to strive to $0$. 
We finally aggregate $\Delta SP(y)$ for all labels using the max operator to get a final scalar metric:

\begin{equation}
\begin{split}
\label{eq:parity_metrics}
\Delta SP = max(\{\Delta SP(y) | y \in \Y\})
\end{split}
\end{equation}

\subsubsection{Performance Metrics}
As our main classification metric, we used the F1 score. We examined both the micro F1 score, which is computed globally based on the true and false predictions, and the macro F1 score, computed per each class and averaged across all classes. For completeness, we also report the Accuracy (ACC).
\section{Experimental Results}
\label{sec:results}
In this section, we report the experimental results. We start by comparing \eqgnn to the baselines (Section~\ref{sec:result_main}). We then demonstrate the importance of $\lambda$ to the \eqgnn architecture (Section~\ref{sec:result_lambda}). We continue by showing the superiority of the permutation loss, compared to other loss functions, both over synthetic datasets (Section~\ref{sec:result_premute_synthetic}) and real datasets (Section~\ref{sec:result_premute_real}). Finally, we explore two qualitative examples, that visualizes the importance of fairness in graphs (Section~\ref{sec:result_qualitative}).

\subsection{Main Result}
\label{sec:result_main}

\begin{table}
\caption{\label{table:main_results} Fairness and performance results}
\center\setlength\tabcolsep{3.0pt}\small
\begin{tabular}{|c|c|cccc|}
\hline
Dataset                         &  Metrics          & GCN            & FairGNN        & Debias          & EqGNN                   \\
\hline 
\multirow{5}{20pt}{Pokec-gender} &  $\Delta EO$ (\%) & $12.3 \pm 1.0$ & $13.7 \pm 1.1$ &  $11.1 \pm 0.7$ & $\mathbf{9.4  \pm 0.8}$ \\
                                &  $\Delta SP$ (\%) & $8.4  \pm 0.6$ & $10.2 \pm 0.4$ &  $8.1  \pm 0.6$ & $\mathbf{4.2  \pm 0.3}$ \\
                                &          ACC (\%) & $65.1 \pm 0.2$ & $65.2 \pm 0.2$ &  $65.0 \pm 0.2$ &         $65.4 \pm 0.2$  \\
                                &     F1-macro (\%) & $61.2 \pm 0.2$ & $61.3 \pm 0.2$ &  $61.1 \pm 0.2$ &         $61.3 \pm 0.2$  \\
                                &     F1-micro (\%) & $66.5 \pm 0.2$ & $66.6 \pm 0.2$ &  $66.5 \pm 0.2$ &         $66.9 \pm 0.2$  \\
\hline 
\multirow{5}{20pt}{Pokec-region} &  $\Delta EO$ (\%) & $17.3 \pm 1.2$ & $17.2 \pm 1.2$ &  $15.8 \pm 1.1$ & $\mathbf{11.2 \pm 1.3}$ \\
                                &  $\Delta SP$ (\%) & $8.4  \pm 0.4$ & $8.2  \pm 0.5$ &  $7.1  \pm 0.4$ & $\mathbf{4.1  \pm 0.5}$ \\
                                &          ACC (\%) & $64.9 \pm 0.2$ & $64.1 \pm 0.3$ &  $63.1 \pm 0.3$ &         $65.1 \pm 0.3$  \\
                                &     F1-macro (\%) & $61.0 \pm 0.2$ & $60.4 \pm 0.2$ &  $59.6 \pm 0.2$ &         $61.3 \pm 0.2$  \\
                                &     F1-micro (\%) & $66.4 \pm 0.2$ & $65.4 \pm 0.3$ &  $64.4 \pm 0.3$ &         $66.5 \pm 0.3$  \\
\hline 
\multirow{5}{20pt}{NBA}          &  $\Delta EO$ (\%) & $25.9 \pm 2.0$ & $24.9 \pm 1.9$ &  $24.0 \pm 1.6$ & $\mathbf{22.1 \pm 2.1}$ \\
                                &  $\Delta SP$ (\%) &  $9.3 \pm 1.4$ &  $8.0 \pm 1.1$ &   $8.0 \pm 1.0$ & $\mathbf{7.0  \pm 1.2}$ \\
                                &          ACC (\%) & $72.9 \pm 0.8$ & $71.6 \pm 1.0$ &  $72.9 \pm 0.7$ &         $72.7 \pm 0.9$  \\
                                &     F1-macro (\%) & $72.7 \pm 0.8$ & $70.6 \pm 1.3$ &  $72.3 \pm 0.7$ &         $72.2 \pm 0.9$  \\
                                &     F1-micro (\%) & $72.8 \pm 0.8$ & $70.8 \pm 1.4$ &  $72.6 \pm 0.7$ &         $72.4 \pm 0.9$  \\
\hline
\end{tabular}
\end{table}
Table~\ref{table:main_results} reports the results of \eqgnn and baselines over the datasets with respect to the performance and fairness metrics.
We can notice that, while the performance metrics are very much similar between all baselines (apart from \debias in Pokec-region), \eqgnn outperforms all other baselines in both fairness metrics. An interesting observation is that \debias is the second best, after \eqgnn, to improve the EO metric, without harming the performance metrics. This can be explained as it is the only baseline to optimize with respect to EO. Additionally, \debias has gained fairness in Pokec-region, but at the cost of performance. This is a general phenomena: the lower the performance, the better the fairness. For example, when the performance is random, surely the algorithm doesn't prefer any particular group and therefore is extremely fair. Here, \eqgnn is able to both optimize the fairness metrics while keeping the performance metrics high. 
The particularly low performance demonstrated by \fairgnn was also validated with the authors of the paper. 
The previously reported results were validated over a single validation step as opposed to several, to insure statistical significance.

\subsection{The Discriminator for Bias Reduction}
\label{sec:result_lambda}

\begin{figure*}
    \center\setlength\tabcolsep{3.0pt}\small
    \includegraphics[width=0.8\textwidth]
    	{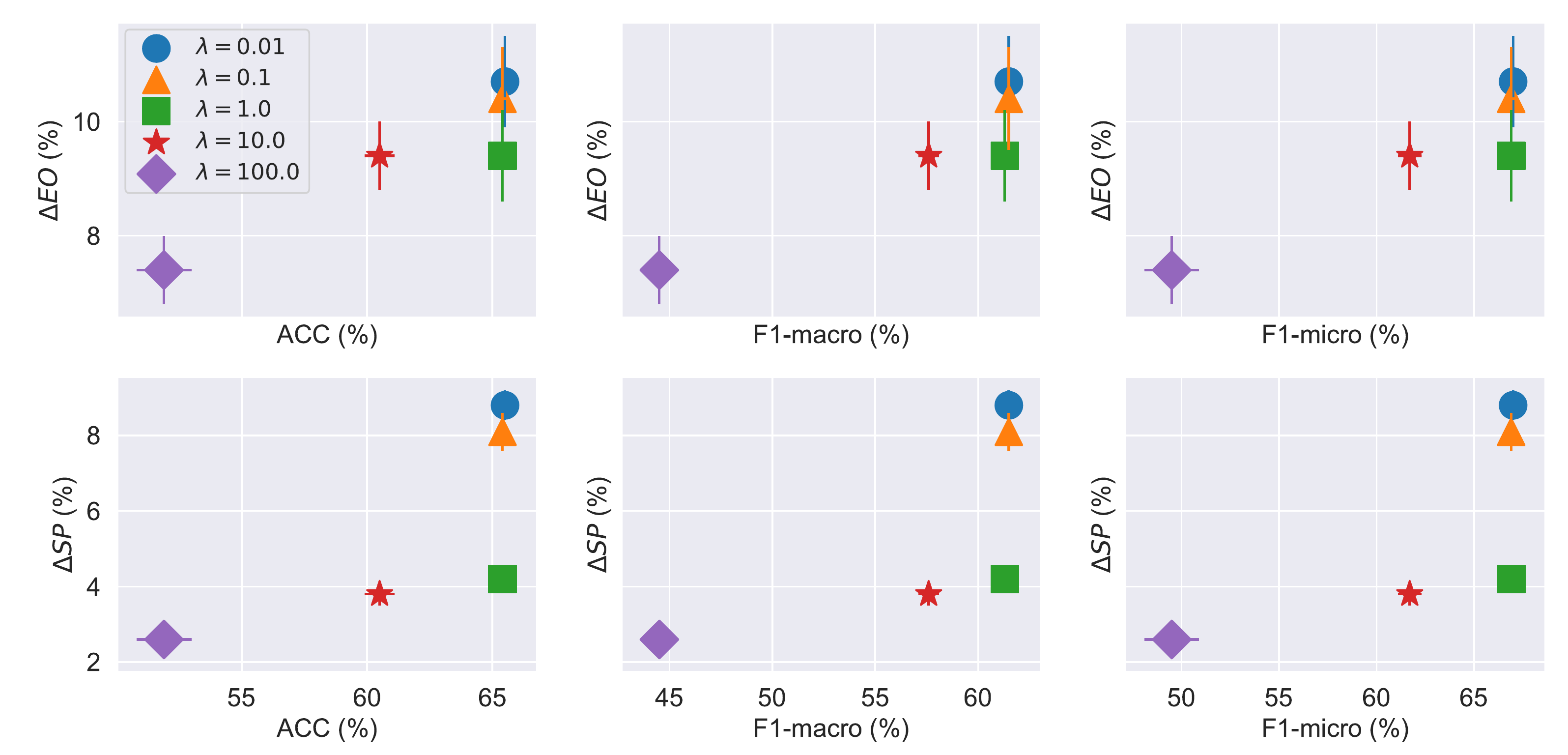}
    % \subfloat[Pokec-gender]
    %     {{\includegraphics[width=0.5\textwidth]
    % 	{lambda-pokec-gender.pdf} }}
    % \quad
    % \vskip\baselineskip
    % \subfloat[Pokec-region]
    %     {{\includegraphics[width=0.5\textwidth]
    % 	{lambda-pokec-region.pdf} }}
    \caption{\label{fig:lambda_results} The comparisons of different $\lambda$ values over the Pokec-gender dataset. Lower-right is better.}
\end{figure*}

As a second analysis, we demonstrate the importance of the $\lambda$ parameter with respect to the performance and fairness metrics. 
The $\lambda$ hyper-parameter serves as a regularization for task performance as opposed to fairness. High values of $\lambda$ cause the discriminator EO regulation on the classifier to be higher.
While \eqgnn results reported in this paper use $\lambda=1$ for Pokec datasets, and $\lambda=0.1$ for NBA, we show additional results for $\lambda \in \{0.01,0.1,1,10,100\}$. In Figure~\ref{fig:lambda_results}, we can observe that the selected $\lambda$s show best results over all metrics for Pokec-gender, while similar results where shown over Pokec-region and NBA. Obviously, enlarging the $\lambda$ results in a more fair model but at the cost of performance. 
% Intuitively, the goal of the \eqgnn architecture, is to find the best curved line between complete performance and complete fairness.
The $\lambda$ hyper-parameter is an issue of priority: depending on the task, one should decide what should be the performance vs. fairness prioritization. Therefore, \eqgnn can be used with any desired $\lambda$ where we chose $\lambda=1$ as it is the elbow of the curve.

\subsection{Synthetic Evaluation of the Permutation Loss}
\label{sec:result_premute_synthetic}

In this experiment, we wish to demonstrate the power of the permutation loss over synthetic data.
Going back to the notations used in Section~\ref{sec:permute}, we generate two paired groups $\X_1, \X_2 \subset \RR^2$ in the following ways:

$\bullet$  \textbf{Rotation}:
$$\X_1 \sim \begin{pmatrix} Z_1 \\ Z_2 \end{pmatrix} \quad,\quad \X_2 \sim \begin{pmatrix} cos \theta & -sin \theta \\ sin \theta & cos \theta \end{pmatrix} \cdot \X_1,$$
where $Z_1,Z_2 \sim \mathcal{N}(0,1)$, and $\theta=\frac{\pi}{2}$.
This can simply be thought as one group is a 2-dimensional Gaussian, while the second is the exact same Gaussian but rotated by $\theta$.
As a rotation over a Gaussian resolves also with a Gaussian, $\X_2$ is also a 2-dimensional Gaussian. Yet, it is paired to $\X_1$, as given a sample from $\X_1$, we can predict its pair from $\X_2$ (simply rotate it by $\theta$).

$\bullet$  \textbf{Shift}:
$$\X_1 \sim \begin{pmatrix} Z_1 \\ Z_2 \end{pmatrix} \quad,\quad \X_2 \sim \X_1 + \begin{pmatrix} \epsilon \\ 0 \end{pmatrix},$$
where $Z_1,Z_2 \sim \mathcal{N}(0,1)$, and $\epsilon=0.1$.
This can simply be thought as one group is a 2-dimensional Gaussian, while the second is the exact same Gaussian but shifted by $\epsilon$ on the first axis.
As shifting a Gaussian by a small value, $\X_2$ overlaps $\X_1$ and therefore, it is hard to distinguish between the two. Yet, it is paired to $\X_1$, as given a sample from $\X_1$, we can predict its pair from $\X_2$ (simply add $\epsilon$).

% While each group separately distributes as a 2-dimensional Gaussian, looking at the pairs, they can easily be discriminated.
% As we use the Permutation Loss as a discriminator, let us from now on call $\X_1$ the real samples, and $\X_2$ the fake samples.

Over these two synthetic datasets, we train four classifiers:

$\bullet$  \textbf{T-test}  - adapted: As the original unpaired t-test requires a one dimensional data, we first map the samples into a single scalar using a fully connected layer and train it using the the t-statistic defined in Section~\ref{def:ttest}.

$\bullet$  \textbf{Paired T-test} - adapted: Similar to \textbf{T-test}, but using the paired t-statistic defined in Section~\ref{def:pairedttest}.

$\bullet$  \textbf{C2ST}~\cite{lopez2016revisiting}: A linear classifier that given a sample, tries to predict to which group it belongs to.

$\bullet$  \textbf{Permutation} (Section~\ref{sec:permute}): A linear classifier that given a randomly shuffled pair, predicts if it was shuffled or not. For detailed implementation please refer to Algorithm~\ref{alg:permute}.

We sample $10000$ pairs for train and an additional $10000$ for test, and average results over 5 runs. In Table \ref{table:loss_results} we report the p-value of the classifiers over the different generated datasets.

\begin{table}
\caption{\label{table:permute_results} P-value comparison of different classifiers over synthetic datasets, lower is better.}
\centering
\begin{tabular}{|l|cc|} 
\hline
\multicolumn{1}{|c|}{Model} & Shift      & Rotation    \\ 
\hline
T-test             & 0.24       & 0.47        \\
Paired T-test      & \textbf{0} & 0.36        \\
C2ST               & 0.5        & 0.5         \\
Permutation        & \textbf{0} & \textbf{0}  \\
\hline
\end{tabular}
\end{table}

We can observe that, the permutation classifier captures the difference between the pairs perfectly in both datasets. This does not hold for the Paired T-test that captures the difference only for the Shift dataset. The reason it classifies well only over the Shift dataset is because it is a linear transformation, which is easier to learn. We can further notice that both unpaired classifiers (T-test and C2ST) perform poorly over both datasets.
% This observation concludes that the permutation classifier significantly outperforms other losses for paired samples and has best results in discriminating between pairs.
% For this reason we chose to use it as a discriminator in our architecture.
The promising results of the permutation classifier on our synthetic datasets, drive us to choose it as the potential discriminator in the \eqgnn architecture.
We validate this choice over real-datasets next section.

\subsection{The Importance of the Permutation Loss}
\label{sec:result_premute_real}

\begin{table}
\caption{\label{table:loss_results} The comparisons of different loss functions.}
\center\setlength\tabcolsep{3.0pt}\small
\begin{tabular}{|c|c|cccc|}
\hline
Dataset                         & Metrics           & Unpaired       & Paired         &  Permutation/h & Permutation        \\
\hline
\multirow{5}{20pt}{Pokec-gender} &  $\Delta EO$ (\%) & $10.9 \pm 1.1$ & $10.5 \pm 0.6$ & $11.4 \pm 0.9$ & $\mathbf{9.4  \pm 0.8}$ \\
                                &  $\Delta SP$ (\%) & $5.8  \pm 0.8$ & $3.9  \pm 0.4$ & $7.2  \pm 1.1$ & $4.2  \pm 0.3$ \\
                                &          ACC (\%) & $65.2 \pm 0.4$ & $62.5 \pm 0.7$ & $66.0 \pm 0.4$ & $65.4 \pm 0.2$ \\
                                &     F1-macro (\%) & $61.1 \pm 0.3$ & $58.6 \pm 0.6$ & $61.7 \pm 0.3$ & $61.3 \pm 0.2$ \\
                                &     F1-micro (\%) & $66.7 \pm 0.4$ & $63.8 \pm 0.7$ & $67.5 \pm 0.3$ & $66.9 \pm 0.2$ \\
\hline 
\multirow{5}{20pt}{Pokec-region} &  $\Delta EO$ (\%) & $15.5 \pm 2.1$ & $13.5 \pm 2.1$ & $16.5 \pm 1.8$ & $\mathbf{11.2 \pm 1.3}$ \\
                                &  $\Delta SP$ (\%) & $6.4  \pm 0.6$ & $4.2  \pm 0.7$ & $7.9  \pm 0.5$ & $4.1  \pm 0.5$ \\
                                &          ACC (\%) & $64.2 \pm 0.4$ & $62.9 \pm 0.5$ & $65.2 \pm 0.2$ & $65.1 \pm 0.3$ \\
                                &     F1-macro (\%) & $60.3 \pm 0.3$ & $59.3 \pm 0.4$ & $61.2 \pm 0.2$ & $61.3 \pm 0.2$ \\
                                &     F1-micro (\%) & $65.6 \pm 0.4$ & $64.2 \pm 0.5$ & $66.7 \pm 0.2$ & $66.5 \pm 0.3$ \\
\hline 
\multirow{5}{20pt}{NBA}          &  $\Delta EO$ (\%) & $22.7 \pm 1.9$ & $25.1 \pm 1.8$ & $22.4 \pm 1.9$ & $\mathbf{22.1 \pm 2.1}$ \\
                                &  $\Delta SP$ (\%) & $7.5  \pm 0.8$ & $7.3  \pm 1.1$ & $7.1  \pm 0.9$ & $7.0  \pm 1.2$ \\
                                &          ACC (\%) & $70.9 \pm 1.1$ & $72.7 \pm 0.7$ & $72.0 \pm 0.9$ & $72.7 \pm 0.9$ \\
                                &     F1-macro (\%) & $68.8 \pm 1.7$ & $71.9 \pm 0.8$ & $70.9 \pm 1.2$ & $72.2 \pm 0.9$ \\
                                &     F1-micro (\%) & $69.2 \pm 1.7$ & $72.2 \pm 0.8$ & $71.2 \pm 1.2$ & $72.4 \pm 0.9$ \\
\hline
\end{tabular}
\end{table}
As an ablation study, we compare different loss functions for the discriminator. We choose to compare the permutation loss with three different loss functions:
(1) Unpaired: Inspired by \cite{romano2020achieving}, an unpaired binary cross-entropy loss as presented in Eq. \ref{eq:adv_classifier}. The loss is estimated by a classifier that predicts if a sample represents a real sensitive attribute or a dummy.
(2) Permutation/h: A permutation loss without concatenating the hidden representation $\h_i$ to the discriminator samples, while leaving the sample to be $(\Y_i, \A_i, \tilde{\A}_i, \hat{\Y}_i)$.
(3) Paired: A paired loss:
\begin{equation}
\begin{split}
\min_{\theta_{\F}}\max_{\theta_{\D}} L_{adv} = \E_{\tilde{\A} \sim \PP_{\A|\Y}(\A|\Y)} [\sigma(\D(\Y  \; || \; \A \; || \; \F(\X,\g) \; || \; \h))
\\
\qquad - \sigma(\D(\Y  \; || \; \tilde{\A} \; || \; \F(\X,\g) \; || \; \h)) ]
\end{split}
\end{equation}
where $\sigma$ is the Sigmoid activation. This loss is the known paired student t-test with a neural network version of it (as demonstrated by \cite{lopez2016revisiting} on the unpaired t-test). Implementation of this loss when summing the absolute differences yielded poor results. We therefore, report a version of this loss with a summation over non absolute differences.
The results of the different loss functions are reported in Table~\ref{table:loss_results}. One can notice that all loss functions have gained fairness over our baselines (as reported in Table~\ref{table:main_results}), while the permutation loss with the hidden representation $\h$, outperforms the others, and specifically Permutation/h. This implies that the hidden representation is important. In the Pokec datasets, the performance metrics are not impacted apart from the paired loss. We hypothesize this is caused due to its non-convexity in adversarial settings. Additionally, the paired loss demonstrates the same phenomena again: the lower the performance, the better the fairness. In the NBA dataset we do not see much difference between the loss functions. This can be explained due to the size of the graph. However, we do see that the permutation loss is the only one to improve fairness metrics while not hurting the performance metrics. Finally, we can notice that, paired loss functions (the permutation loss and the paired loss) perform better than the unpaired loss (apart from NBA, where the unpaired loss hurts the performance metrics). This can be explained by our paired problem, where we check for the difference between two scenarios of a node (real and fake). This illustrates the general importance of a paired loss function for paired problems.

% \iffalse
\subsection{Qualitative Example}
\label{sec:result_qualitative}

\begin{figure}
    \centering
    \subfloat[Male]
        {
        \label{fig:qualitative_male}
        \includegraphics[width=0.4\textwidth]{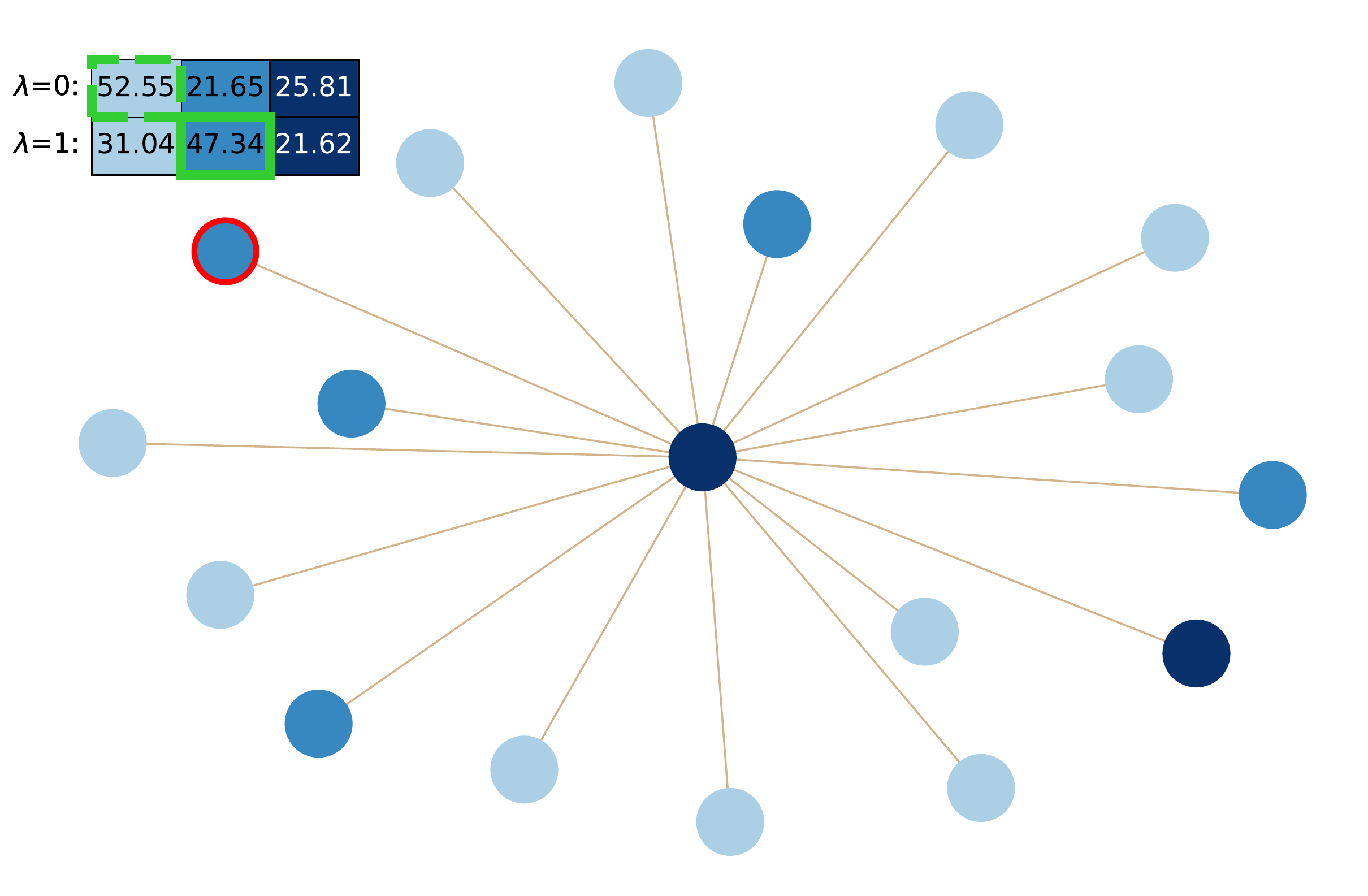}
        }
    \quad
    \subfloat[Female]
    	{
    	\label{fig:qualitative_female}
    	\includegraphics[width=0.4\textwidth]{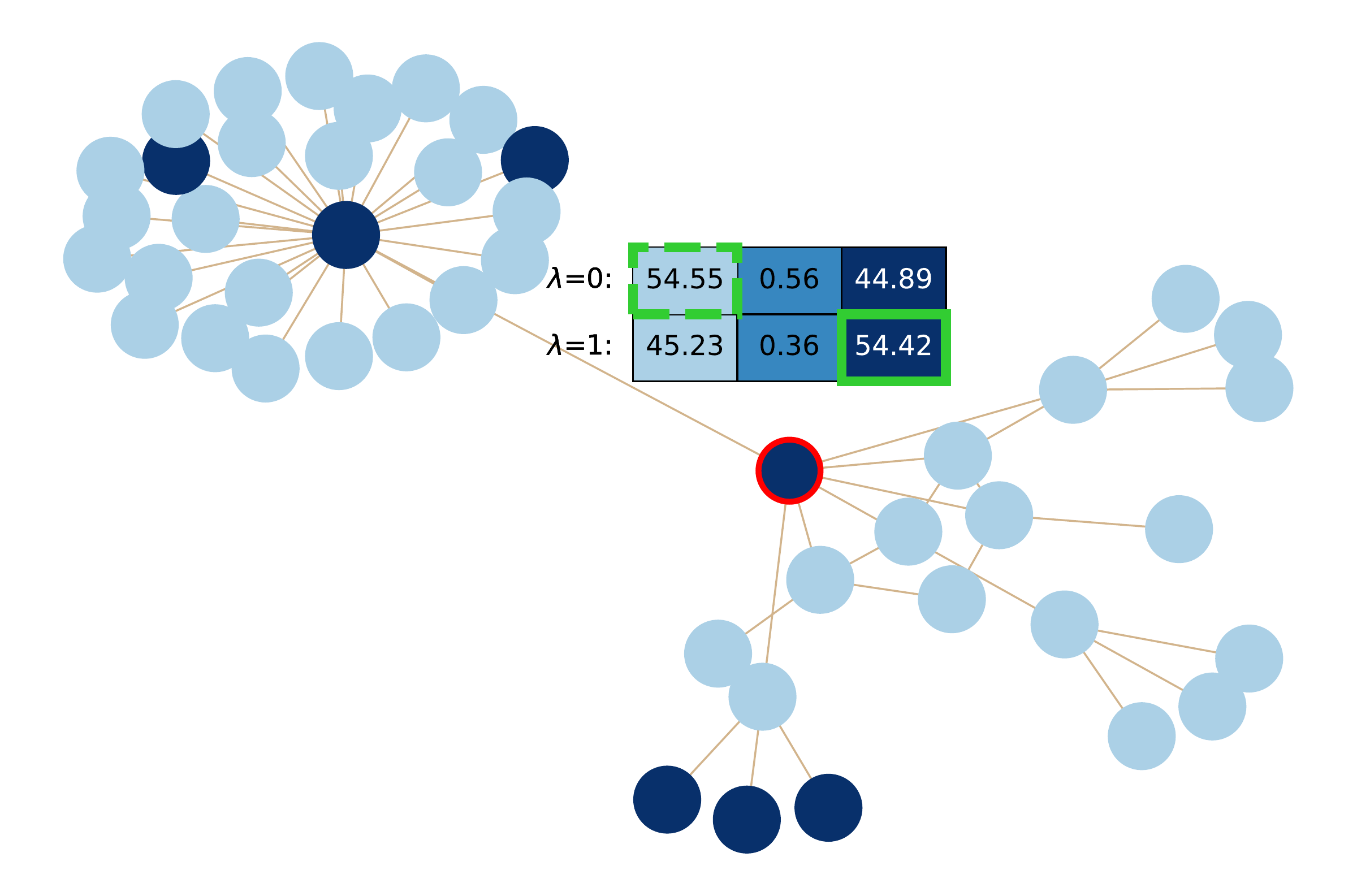}
    	}
    % \quad
    % \vskip\baselineskip
    \caption{\label{fig:qualitative}In each of the 2-hop sub-graphs there is a central node (highlighted in red). The node (and over $80\%$ of his neighbors) are from a specific gender. The node colors represent the class they belong to. Above the central node we can observe the output of the classifier, representing the probabilities of belonging to each class, both for $\lambda=0$ and $\lambda=1$.
    % Notice that $\lambda=1$ predicts the correct class (surrounded with a green rectangular) rather then the predicting the neighbors class, as done by $\lambda=0$ (surrounded with the a dashed rectangular).
    }
\end{figure}

We end this section with a few qualitative examples over the Pokec-gender test-set.
Specifically, we present two qualitative examples, where the central node has the same sensitive attribute as over $80\%$ of its 2-hop neighbors, but holds a different label from most of them. 
We consider 2-hops, as our classifier includes 2 GCN layers.
Figure~\ref{fig:qualitative_male} presents the example for a sensitive attribute being a male where in Figure~\ref{fig:qualitative_female} being a female; i.e., nodes in Figure~\ref{fig:qualitative_male} are mostly males and in Figure~\ref{fig:qualitative_female} mostly females.
A biased approach would be to be inclined to predict the same label for the central node as its same-gender neighbors.
% After further filtering all test-nodes with at least five neighbors, we ended up with only two relevant nodes (one male and one female).
Above the central node we observe the prediction distribution for $\lambda=0$ and $\lambda=1$. 
In Figure~\ref{fig:qualitative_male} we observe that, when no discriminator is applied ($\lambda=0$), and therefore there is no regularization for bias, the probability is $52.55\%$ towards most neighbors label. On the other hand, when applying the discriminator ($\lambda=1$), the probability for that class drops to $31.04\%$, and rises to $47.34\%$ for the correct label. 
Similarly, in Figure~\ref{fig:qualitative_female}, we observe that, when no discriminator is applied ($\lambda=0$), the probability is $54.55\%$ towards most neighbors label. Again, this in comparison to the case when applying the discriminator ($\lambda=1$), the probability for that class drops to $45.23\%$, and rises to $54.42\%$ for the correct label. 
% We can observe the first-hop of of the two nodes. We can notice that for $\lambda=0$ the predictions are closer to their neighbors, while for $\lambda=1$ the predictions are closer to the real label of the node.
These qualitative examples show that, a \emph{equalized odds} regulator over graphs can help make less biased predictions, even when the neighbours at the graphs might cause bias.
% \fi
\section{Conclusions}
\label{sec:conclusions}
In this work, we explored fairness in graphs.
Unlike previous work which optimize the \emph{statistical parity} (SP) fairness criterion, we present a method that learns to optimize \emph{equalized odds} (EO). While SP promises equal chances between groups, it might cripple the utility of the prediction task as it does not give equalized opportunity as EO.
We propose a method that trains a GNN model to neglect information regarding the sensitive attribute only with respect to its label. 
Our method pretrains a sampler to learn the distribution of the sensitive attributes of the nodes given their labels. We then continue training a GNN classifier while regularizing it with a discriminator that discriminates between true and sampled sensitive attributes using a novel loss function -- the ``permutation loss''. This loss allows comparison of pairs. For the unique setup of adversarial learning over graphs, we show it brings performance gains both in utility and in EO. 
While this work uses the loss for the specific case of nodes in two scenarios: fake and true, this loss is general and can be used for any paired problem. 

For future work, we wish to test the novel loss over additional architectures and tasks.
We draw the reader attention that the C2ST discriminator is the commonly used discriminator for many architectures that work over paired data. For instance, the pix2pix architecture \cite{isola2017image} is a classic architecture that inspired many works. Although the pix2pix discriminator receives paired samples, it is still just an advanced C2ST discriminator. Alternatively, using a paired discriminator instead, can create a much powerful discriminator and therefore a much powerful generator. Observing many works that apply over paired samples, we haven't found any architectures that are designed to work over paired samples.
We believe that, although this work uses the permutation loss for a specific use-case, it is a general architecture that can be used for any paired problem. 

We empirically show that our method outperforms different baselines in the combined fairness-performance metrics, over datasets with different attributes and sizes.
To the best of our knowledge, we are the first to optimize GNNs for the EO criteria and hope it will serve as a beacon for works to come.

\bibliographystyle{ACM-Reference-Format}
\bibliography{eqgnn} 


\begin{thebibliography}{36}

%%% the Web address.

\ifx \showCODEN    \undefined \def \showCODEN     #1{\unskip}     \fi
\ifx \showDOI      \undefined \def \showDOI       #1{#1}\fi
\ifx \showISBNx    \undefined \def \showISBNx     #1{\unskip}     \fi
\ifx \showISBNxiii \undefined \def \showISBNxiii  #1{\unskip}     \fi
\ifx \showISSN     \undefined \def \showISSN      #1{\unskip}     \fi
\ifx \showLCCN     \undefined \def \showLCCN      #1{\unskip}     \fi
\ifx \shownote     \undefined \def \shownote      #1{#1}          \fi
\ifx \showarticletitle \undefined \def \showarticletitle #1{#1}   \fi
\ifx \showURL      \undefined \def \showURL       {\relax}        \fi
% invisible to TeX
\providecommand\bibfield[2]{#2}
\providecommand\bibinfo[2]{#2}
\providecommand\natexlab[1]{#1}
\providecommand\showeprint[2][]{arXiv:#2}

\bibitem[\protect\citeauthoryear{Barocas, Hardt, and Narayanan}{Barocas
  et~al\mbox{.}}{2019}]%
        {barocas-hardt-narayanan}
\bibfield{author}{\bibinfo{person}{Solon Barocas}, \bibinfo{person}{Moritz
  Hardt}, {and} \bibinfo{person}{Arvind Narayanan}.}
  \bibinfo{year}{2019}\natexlab{}.
\newblock \bibinfo{booktitle}{\emph{Fairness and Machine Learning}}.
\newblock \bibinfo{publisher}{fairmlbook.org}.
\newblock
\newblock
\shownote{\url{http://www.fairmlbook.org}.}


\bibitem[\protect\citeauthoryear{Belkin and Niyogi}{Belkin and Niyogi}{2001}]%
        {Laplacian:NIPS2001}
\bibfield{author}{\bibinfo{person}{Mikhail Belkin} {and}
  \bibinfo{person}{Partha Niyogi}.} \bibinfo{year}{2001}\natexlab{}.
\newblock \showarticletitle{Laplacian eigenmaps and spectral techniques for
  embedding and clustering.}. In \bibinfo{booktitle}{\emph{Advances in Neural
  Information Processing Systems}}, Vol.~\bibinfo{volume}{14}.
  \bibinfo{pages}{585--591}.
\newblock


\bibitem[\protect\citeauthoryear{Biewald}{Biewald}{2020}]%
        {wandb}
\bibfield{author}{\bibinfo{person}{Lukas Biewald}.}
  \bibinfo{year}{2020}\natexlab{}.
\newblock \bibinfo{title}{Experiment Tracking with Weights and Biases}.
\newblock
\newblock
\urldef\tempurl%
\url{https://www.wandb.com/}
\showURL{%
\tempurl}
\newblock
\shownote{Software available from wandb.com.}


\bibitem[\protect\citeauthoryear{Bose and Hamilton}{Bose and Hamilton}{2019}]%
        {bose2019compositional}
\bibfield{author}{\bibinfo{person}{Avishek Bose} {and} \bibinfo{person}{William
  Hamilton}.} \bibinfo{year}{2019}\natexlab{}.
\newblock \showarticletitle{Compositional fairness constraints for graph
  embeddings}. In \bibinfo{booktitle}{\emph{International Conference on Machine
  Learning}}. PMLR, \bibinfo{pages}{715--724}.
\newblock


\bibitem[\protect\citeauthoryear{Buyl and De~Bie}{Buyl and De~Bie}{2020}]%
        {buyl2020debayes}
\bibfield{author}{\bibinfo{person}{Maarten Buyl} {and} \bibinfo{person}{Tijl
  De~Bie}.} \bibinfo{year}{2020}\natexlab{}.
\newblock \showarticletitle{DeBayes: a Bayesian method for debiasing network
  embeddings}. In \bibinfo{booktitle}{\emph{International Conference on Machine
  Learning}}. PMLR, \bibinfo{pages}{1220--1229}.
\newblock


\bibitem[\protect\citeauthoryear{Cui, Henrickson, Ke, and Wang}{Cui
  et~al\mbox{.}}{2019}]%
        {cui2019traffic}
\bibfield{author}{\bibinfo{person}{Zhiyong Cui}, \bibinfo{person}{Kristian
  Henrickson}, \bibinfo{person}{Ruimin Ke}, {and} \bibinfo{person}{Yinhai
  Wang}.} \bibinfo{year}{2019}\natexlab{}.
\newblock \showarticletitle{Traffic graph convolutional recurrent neural
  network: A deep learning framework for network-scale traffic learning and
  forecasting}.
\newblock \bibinfo{journal}{\emph{IEEE Transactions on Intelligent
  Transportation Systems}} (\bibinfo{year}{2019}).
\newblock


\bibitem[\protect\citeauthoryear{Dai and Wang}{Dai and Wang}{2021}]%
        {dai2020say}
\bibfield{author}{\bibinfo{person}{Enyan Dai} {and} \bibinfo{person}{Suhang
  Wang}.} \bibinfo{year}{2021}\natexlab{}.
\newblock \showarticletitle{Say No to the Discrimination: Learning Fair Graph
  Neural Networks with Limited Sensitive Attribute Information}. In
  \bibinfo{booktitle}{\emph{Proceedings of the 14th ACM International
  Conference on Web Search and Data Mining}}. \bibinfo{pages}{680--688}.
\newblock


\bibitem[\protect\citeauthoryear{Dwork, Hardt, Pitassi, Reingold, and
  Zemel}{Dwork et~al\mbox{.}}{2012}]%
        {dwork2012fairness}
\bibfield{author}{\bibinfo{person}{Cynthia Dwork}, \bibinfo{person}{Moritz
  Hardt}, \bibinfo{person}{Toniann Pitassi}, \bibinfo{person}{Omer Reingold},
  {and} \bibinfo{person}{Richard Zemel}.} \bibinfo{year}{2012}\natexlab{}.
\newblock \showarticletitle{Fairness through awareness}. In
  \bibinfo{booktitle}{\emph{Proceedings of the 3rd innovations in theoretical
  computer science conference}}. \bibinfo{pages}{214--226}.
\newblock


\bibitem[\protect\citeauthoryear{Falcon}{Falcon}{2019}]%
        {falcon2019pytorch}
\bibfield{author}{\bibinfo{person}{WA Falcon}.}
  \bibinfo{year}{2019}\natexlab{}.
\newblock \showarticletitle{PyTorch Lightning}.
\newblock \bibinfo{journal}{\emph{GitHub. Note:
  https://github.com/PyTorchLightning/pytorch-lightning}}  \bibinfo{volume}{3}
  (\bibinfo{year}{2019}).
\newblock


\bibitem[\protect\citeauthoryear{Grover and Leskovec}{Grover and
  Leskovec}{2016}]%
        {grover16node}
\bibfield{author}{\bibinfo{person}{Aditya Grover} {and} \bibinfo{person}{Jure
  Leskovec}.} \bibinfo{year}{2016}\natexlab{}.
\newblock \showarticletitle{node2vec: Scalable feature learning for networks}.
  In \bibinfo{booktitle}{\emph{Proceedings of the 22nd ACM SIGKDD international
  conference on Knowledge discovery and data mining}}.
  \bibinfo{pages}{855--864}.
\newblock


\bibitem[\protect\citeauthoryear{Hamilton, Ying, and Leskovec}{Hamilton
  et~al\mbox{.}}{2017}]%
        {hamilton2017inductive}
\bibfield{author}{\bibinfo{person}{William~L Hamilton}, \bibinfo{person}{Rex
  Ying}, {and} \bibinfo{person}{Jure Leskovec}.}
  \bibinfo{year}{2017}\natexlab{}.
\newblock \showarticletitle{Inductive representation learning on large graphs}.
  In \bibinfo{booktitle}{\emph{Proceedings of the 31st International Conference
  on Neural Information Processing Systems}}. \bibinfo{pages}{1025--1035}.
\newblock


\bibitem[\protect\citeauthoryear{Hardt, Price, and Srebro}{Hardt
  et~al\mbox{.}}{2016}]%
        {hardt2016equality}
\bibfield{author}{\bibinfo{person}{Moritz Hardt}, \bibinfo{person}{Eric Price},
  {and} \bibinfo{person}{Nati Srebro}.} \bibinfo{year}{2016}\natexlab{}.
\newblock \showarticletitle{Equality of Opportunity in Supervised Learning}.
\newblock \bibinfo{journal}{\emph{Advances in Neural Information Processing
  Systems}}  \bibinfo{volume}{29} (\bibinfo{year}{2016}),
  \bibinfo{pages}{3315--3323}.
\newblock


\bibitem[\protect\citeauthoryear{Isola, Zhu, Zhou, and Efros}{Isola
  et~al\mbox{.}}{2017}]%
        {isola2017image}
\bibfield{author}{\bibinfo{person}{Phillip Isola}, \bibinfo{person}{Jun-Yan
  Zhu}, \bibinfo{person}{Tinghui Zhou}, {and} \bibinfo{person}{Alexei~A
  Efros}.} \bibinfo{year}{2017}\natexlab{}.
\newblock \showarticletitle{Image-to-image translation with conditional
  adversarial networks}. In \bibinfo{booktitle}{\emph{Proceedings of the IEEE
  conference on computer vision and pattern recognition}}.
  \bibinfo{pages}{1125--1134}.
\newblock


\bibitem[\protect\citeauthoryear{Kang, Lijffijt, and Bie}{Kang
  et~al\mbox{.}}{2019}]%
        {Kang2019ConditionalNE}
\bibfield{author}{\bibinfo{person}{Bo Kang}, \bibinfo{person}{Jefrey Lijffijt},
  {and} \bibinfo{person}{T.~D. Bie}.} \bibinfo{year}{2019}\natexlab{}.
\newblock \showarticletitle{Conditional Network Embeddings}.
\newblock \bibinfo{journal}{\emph{International Conference on Learning
  Representations}} (\bibinfo{year}{2019}).
\newblock


\bibitem[\protect\citeauthoryear{Kang, He, Maciejewski, and Tong}{Kang
  et~al\mbox{.}}{2020}]%
        {kang2020inform}
\bibfield{author}{\bibinfo{person}{Jian Kang}, \bibinfo{person}{Jingrui He},
  \bibinfo{person}{Ross Maciejewski}, {and} \bibinfo{person}{Hanghang Tong}.}
  \bibinfo{year}{2020}\natexlab{}.
\newblock \showarticletitle{InFoRM: Individual Fairness on Graph Mining}. In
  \bibinfo{booktitle}{\emph{Proceedings of the 26th ACM SIGKDD International
  Conference on Knowledge Discovery \& Data Mining}}.
  \bibinfo{pages}{379--389}.
\newblock


\bibitem[\protect\citeauthoryear{Kingma and Ba}{Kingma and Ba}{2015}]%
        {kingma2014adam}
\bibfield{author}{\bibinfo{person}{Diederik~P Kingma} {and}
  \bibinfo{person}{Jimmy Ba}.} \bibinfo{year}{2015}\natexlab{}.
\newblock \showarticletitle{Adam: A method for stochastic optimization}.
\newblock \bibinfo{journal}{\emph{International Conference on Learning
  Representations}} (\bibinfo{year}{2015}).
\newblock


\bibitem[\protect\citeauthoryear{Kipf and Welling}{Kipf and Welling}{2017}]%
        {kipf2017semi}
\bibfield{author}{\bibinfo{person}{Thomas~N. Kipf} {and} \bibinfo{person}{Max
  Welling}.} \bibinfo{year}{2017}\natexlab{}.
\newblock \showarticletitle{Semi-Supervised Classification with Graph
  Convolutional Networks}.
\newblock \bibinfo{journal}{\emph{International Conference on Learning
  Representations}} (\bibinfo{year}{2017}).
\newblock


\bibitem[\protect\citeauthoryear{Lopez-Paz and Oquab}{Lopez-Paz and
  Oquab}{2017}]%
        {lopez2016revisiting}
\bibfield{author}{\bibinfo{person}{David Lopez-Paz} {and}
  \bibinfo{person}{Maxime Oquab}.} \bibinfo{year}{2017}\natexlab{}.
\newblock \showarticletitle{Revisiting classifier two-sample tests}. In
  \bibinfo{booktitle}{\emph{International Conference on Learning
  Representations}}.
\newblock


\bibitem[\protect\citeauthoryear{Obermeyer, Powers, Vogeli, and
  Mullainathan}{Obermeyer et~al\mbox{.}}{2019}]%
        {obermeyer2019dissecting}
\bibfield{author}{\bibinfo{person}{Ziad Obermeyer}, \bibinfo{person}{Brian
  Powers}, \bibinfo{person}{Christine Vogeli}, {and} \bibinfo{person}{Sendhil
  Mullainathan}.} \bibinfo{year}{2019}\natexlab{}.
\newblock \showarticletitle{Dissecting racial bias in an algorithm used to
  manage the health of populations}.
\newblock \bibinfo{journal}{\emph{Science}} \bibinfo{volume}{366},
  \bibinfo{number}{6464} (\bibinfo{year}{2019}), \bibinfo{pages}{447--453}.
\newblock


\bibitem[\protect\citeauthoryear{Od{\'e}n, Wedel, et~al\mbox{.}}{Od{\'e}n
  et~al\mbox{.}}{1975}]%
        {oden1975arguments}
\bibfield{author}{\bibinfo{person}{Anders Od{\'e}n}, \bibinfo{person}{Hans
  Wedel}, {et~al\mbox{.}}} \bibinfo{year}{1975}\natexlab{}.
\newblock \showarticletitle{Arguments for Fisher's permutation test}.
\newblock \bibinfo{journal}{\emph{The Annals of Statistics}}
  \bibinfo{volume}{3}, \bibinfo{number}{2} (\bibinfo{year}{1975}),
  \bibinfo{pages}{518--520}.
\newblock


\bibitem[\protect\citeauthoryear{Pedreshi, Ruggieri, and Turini}{Pedreshi
  et~al\mbox{.}}{2008}]%
        {pedreshi2008discrimination}
\bibfield{author}{\bibinfo{person}{Dino Pedreshi}, \bibinfo{person}{Salvatore
  Ruggieri}, {and} \bibinfo{person}{Franco Turini}.}
  \bibinfo{year}{2008}\natexlab{}.
\newblock \showarticletitle{Discrimination-aware data mining}. In
  \bibinfo{booktitle}{\emph{Proceedings of the 14th ACM SIGKDD international
  conference on Knowledge discovery and data mining}}.
  \bibinfo{pages}{560--568}.
\newblock


\bibitem[\protect\citeauthoryear{Perozzi, Al-Rfou, and Skiena}{Perozzi
  et~al\mbox{.}}{2014}]%
        {Perozzi:2014:DOL}
\bibfield{author}{\bibinfo{person}{Bryan Perozzi}, \bibinfo{person}{Rami
  Al-Rfou}, {and} \bibinfo{person}{Steven Skiena}.}
  \bibinfo{year}{2014}\natexlab{}.
\newblock \showarticletitle{Deepwalk: Online learning of social
  representations}. In \bibinfo{booktitle}{\emph{Proceedings of the 20th ACM
  SIGKDD international conference on Knowledge discovery and data mining}}.
  \bibinfo{pages}{701--710}.
\newblock


\bibitem[\protect\citeauthoryear{Rahman, Surma, Backes, and Zhang}{Rahman
  et~al\mbox{.}}{2019}]%
        {rahman2019fairwalk}
\bibfield{author}{\bibinfo{person}{Tahleen~A Rahman},
  \bibinfo{person}{Bartlomiej Surma}, \bibinfo{person}{Michael Backes}, {and}
  \bibinfo{person}{Yang Zhang}.} \bibinfo{year}{2019}\natexlab{}.
\newblock \showarticletitle{Fairwalk: Towards Fair Graph Embedding.}. In
  \bibinfo{booktitle}{\emph{Proceedings of the Twenty-Eighth International
  Joint Conference on Artificial Intelligence, {IJCAI-19}}}.
  \bibinfo{pages}{3289--3295}.
\newblock


\bibitem[\protect\citeauthoryear{Romano, Bates, and Cand{\`e}s}{Romano
  et~al\mbox{.}}{2020}]%
        {romano2020achieving}
\bibfield{author}{\bibinfo{person}{Yaniv Romano}, \bibinfo{person}{Stephen
  Bates}, {and} \bibinfo{person}{Emmanuel~J Cand{\`e}s}.}
  \bibinfo{year}{2020}\natexlab{}.
\newblock \showarticletitle{Achieving Equalized Odds by Resampling Sensitive
  Attributes}.
\newblock \bibinfo{journal}{\emph{Advances in Neural Information Processing
  Systems}} (\bibinfo{year}{2020}).
\newblock


\bibitem[\protect\citeauthoryear{Roweis and Saul}{Roweis and Saul}{2000}]%
        {roweis2000nonlinear}
\bibfield{author}{\bibinfo{person}{Sam~T Roweis} {and}
  \bibinfo{person}{Lawrence~K Saul}.} \bibinfo{year}{2000}\natexlab{}.
\newblock \showarticletitle{Nonlinear dimensionality reduction by locally
  linear embedding}.
\newblock \bibinfo{journal}{\emph{science}} \bibinfo{volume}{290},
  \bibinfo{number}{5500} (\bibinfo{year}{2000}), \bibinfo{pages}{2323--2326}.
\newblock


\bibitem[\protect\citeauthoryear{Singer, Guy, and Radinsky}{Singer
  et~al\mbox{.}}{2019}]%
        {singer2019node}
\bibfield{author}{\bibinfo{person}{Uriel Singer}, \bibinfo{person}{Ido Guy},
  {and} \bibinfo{person}{Kira Radinsky}.} \bibinfo{year}{2019}\natexlab{}.
\newblock \showarticletitle{Node Embedding over Temporal Graphs}. In
  \bibinfo{booktitle}{\emph{Proceedings of the Twenty-Eighth International
  Joint Conference on Artificial Intelligence, {IJCAI-19}}}.
  \bibinfo{publisher}{International Joint Conferences on Artificial
  Intelligence Organization}, \bibinfo{pages}{4605--4612}.
\newblock
\urldef\tempurl%
\url{https://doi.org/10.24963/ijcai.2019/640}
\showDOI{\tempurl}


\bibitem[\protect\citeauthoryear{Singer, Radinsky, and Horvitz}{Singer
  et~al\mbox{.}}{2020}]%
        {10.1093/bioinformatics/btaa1036}
\bibfield{author}{\bibinfo{person}{Uriel Singer}, \bibinfo{person}{Kira
  Radinsky}, {and} \bibinfo{person}{Eric Horvitz}.}
  \bibinfo{year}{2020}\natexlab{}.
\newblock \showarticletitle{{On biases of attention in scientific discovery}}.
\newblock \bibinfo{journal}{\emph{Bioinformatics}} (\bibinfo{date}{12}
  \bibinfo{year}{2020}).
\newblock
\showISSN{1367-4803}
\urldef\tempurl%
\url{https://doi.org/10.1093/bioinformatics/btaa1036}
\showDOI{\tempurl}
\newblock
\shownote{btaa1036.}


\bibitem[\protect\citeauthoryear{Takac and Zabovsky}{Takac and
  Zabovsky}{2012}]%
        {takac2012data}
\bibfield{author}{\bibinfo{person}{Lubos Takac} {and} \bibinfo{person}{Michal
  Zabovsky}.} \bibinfo{year}{2012}\natexlab{}.
\newblock \showarticletitle{Data analysis in public social networks}. In
  \bibinfo{booktitle}{\emph{International scientific conference and
  international workshop present day trends of innovations}},
  Vol.~\bibinfo{volume}{1}.
\newblock


\bibitem[\protect\citeauthoryear{Tenenbaum, De~Silva, and Langford}{Tenenbaum
  et~al\mbox{.}}{2000}]%
        {tenenbaum:global:2000}
\bibfield{author}{\bibinfo{person}{Joshua~B Tenenbaum}, \bibinfo{person}{Vin
  De~Silva}, {and} \bibinfo{person}{John~C Langford}.}
  \bibinfo{year}{2000}\natexlab{}.
\newblock \showarticletitle{A global geometric framework for nonlinear
  dimensionality reduction}.
\newblock \bibinfo{journal}{\emph{science}} \bibinfo{volume}{290},
  \bibinfo{number}{5500} (\bibinfo{year}{2000}), \bibinfo{pages}{2319--2323}.
\newblock


\bibitem[\protect\citeauthoryear{Veli{\v{c}}kovi{\'c}, Cucurull, Casanova,
  Romero, Lio, and Bengio}{Veli{\v{c}}kovi{\'c} et~al\mbox{.}}{2017}]%
        {velivckovic2017graph}
\bibfield{author}{\bibinfo{person}{Petar Veli{\v{c}}kovi{\'c}},
  \bibinfo{person}{Guillem Cucurull}, \bibinfo{person}{Arantxa Casanova},
  \bibinfo{person}{Adriana Romero}, \bibinfo{person}{Pietro Lio}, {and}
  \bibinfo{person}{Yoshua Bengio}.} \bibinfo{year}{2017}\natexlab{}.
\newblock \showarticletitle{Graph attention networks}.
\newblock \bibinfo{journal}{\emph{International Conference on Learning
  Representations}} (\bibinfo{year}{2017}).
\newblock


\bibitem[\protect\citeauthoryear{Wang and Li}{Wang and Li}{2017}]%
        {Wang:2017:RRL}
\bibfield{author}{\bibinfo{person}{Hongjian Wang} {and}
  \bibinfo{person}{Zhenhui Li}.} \bibinfo{year}{2017}\natexlab{}.
\newblock \showarticletitle{Region Representation Learning via Mobility Flow}.
  In \bibinfo{booktitle}{\emph{Proceedings of the 2017 ACM on Conference on
  Information and Knowledge Management}}. \bibinfo{pages}{237--246}.
\newblock


\bibitem[\protect\citeauthoryear{Wu, Pan, Chen, Long, Zhang, and Philip}{Wu
  et~al\mbox{.}}{2020}]%
        {wu2020comprehensive}
\bibfield{author}{\bibinfo{person}{Zonghan Wu}, \bibinfo{person}{Shirui Pan},
  \bibinfo{person}{Fengwen Chen}, \bibinfo{person}{Guodong Long},
  \bibinfo{person}{Chengqi Zhang}, {and} \bibinfo{person}{S~Yu Philip}.}
  \bibinfo{year}{2020}\natexlab{}.
\newblock \showarticletitle{A comprehensive survey on graph neural networks}.
\newblock \bibinfo{journal}{\emph{IEEE Transactions on Neural Networks and
  Learning Systems}} (\bibinfo{year}{2020}).
\newblock


\bibitem[\protect\citeauthoryear{Yan, Xiong, and Lin}{Yan
  et~al\mbox{.}}{2018}]%
        {stgcn2018aaai}
\bibfield{author}{\bibinfo{person}{Sijie Yan}, \bibinfo{person}{Yuanjun Xiong},
  {and} \bibinfo{person}{Dahua Lin}.} \bibinfo{year}{2018}\natexlab{}.
\newblock \showarticletitle{Spatial Temporal Graph Convolutional Networks for
  Skeleton-Based Action Recognition}. In \bibinfo{booktitle}{\emph{Proceedings
  of the AAAI conference on artificial intelligence}},
  Vol.~\bibinfo{volume}{32}.
\newblock


\bibitem[\protect\citeauthoryear{Yan, Xu, Zhang, Zhang, Yang, and Lin}{Yan
  et~al\mbox{.}}{2006}]%
        {Yan:2007:GEE}
\bibfield{author}{\bibinfo{person}{Shuicheng Yan}, \bibinfo{person}{Dong Xu},
  \bibinfo{person}{Benyu Zhang}, \bibinfo{person}{Hong-Jiang Zhang},
  \bibinfo{person}{Qiang Yang}, {and} \bibinfo{person}{Stephen Lin}.}
  \bibinfo{year}{2006}\natexlab{}.
\newblock \showarticletitle{Graph embedding and extensions: A general framework
  for dimensionality reduction}.
\newblock \bibinfo{journal}{\emph{IEEE transactions on pattern analysis and
  machine intelligence}} \bibinfo{volume}{29}, \bibinfo{number}{1}
  (\bibinfo{year}{2006}), \bibinfo{pages}{40--51}.
\newblock


\bibitem[\protect\citeauthoryear{Yu, Yin, and Zhu}{Yu et~al\mbox{.}}{2018}]%
        {yu2018spatio}
\bibfield{author}{\bibinfo{person}{Bing Yu}, \bibinfo{person}{Haoteng Yin},
  {and} \bibinfo{person}{Zhanxing Zhu}.} \bibinfo{year}{2018}\natexlab{}.
\newblock \showarticletitle{Spatio-temporal graph convolutional networks: a
  deep learning framework for traffic forecasting}. In
  \bibinfo{booktitle}{\emph{Proceedings of the 27th International Joint
  Conference on Artificial Intelligence}}. \bibinfo{pages}{3634--3640}.
\newblock


\bibitem[\protect\citeauthoryear{Zhang, Lemoine, and Mitchell}{Zhang
  et~al\mbox{.}}{2018}]%
        {zhang2018mitigating}
\bibfield{author}{\bibinfo{person}{Brian~Hu Zhang}, \bibinfo{person}{Blake
  Lemoine}, {and} \bibinfo{person}{Margaret Mitchell}.}
  \bibinfo{year}{2018}\natexlab{}.
\newblock \showarticletitle{Mitigating unwanted biases with adversarial
  learning}. In \bibinfo{booktitle}{\emph{Proceedings of the 2018 AAAI/ACM
  Conference on AI, Ethics, and Society}}. \bibinfo{pages}{335--340}.
\newblock


\end{thebibliography}

\end{document}